\documentclass[10pt,twocolumn,letterpaper]{article}

\usepackage{iccv}
\usepackage{times}
\usepackage{epsfig}
\usepackage{graphicx}
\usepackage{amsmath}
\usepackage{amssymb}
\usepackage{multirow}
\usepackage{pifont}
\usepackage{booktabs}
\usepackage{epsfig}
\usepackage{booktabs,multirow}
\usepackage{graphics}
\usepackage{threeparttable}
\usepackage{color}
\usepackage[normalem]{ulem}
\usepackage{multirow}
\usepackage{float}
\usepackage{amsfonts}
\usepackage{bm}
\usepackage{enumitem}
\usepackage[numbers]{natbib}
\usepackage{array}
\usepackage[table]{xcolor}
\usepackage{colortbl}
\usepackage{ulem}
\definecolor{myy}{RGB}{126,95,0}
\definecolor{mygray}{gray}{.9}
\definecolor{bblue}{RGB}{30,80,120}
\definecolor{mygray1}{gray}{.7}
\usepackage{mathtools}
\usepackage{subcaption}
\usepackage{siunitx}
\usepackage[nopar]{lipsum}
\usepackage[export]{adjustbox}

\definecolor{mygreen}{HTML}{39b54a}  

\makeatletter
\newcommand{\thickhline}{%
	\noalign {\ifnum 0=`}\fi \hrule height 1pt
	\futurelet \reserved@a \@xhline
}

\newcommand{\core}{\textbf{\textsc{Core}}\xspace}

\usepackage[pagebackref=true,breaklinks=true,letterpaper=true,colorlinks,bookmarks=false]{hyperref}

\iccvfinalcopy 

\def\iccvPaperID{****} 

\ificcvfinal\fi

\begin{document}

\title{\core: Cooperative Reconstruction for Multi-Agent Perception}

\author{Binglu Wang$^{1,2}$\thanks{The first two authors contribute equally to this work.}~, Lei Zhang$^{1*}$,~~Zhaozhong Wang$^{1}$,~~Yongqiang Zhao$^{1}$,~~Tianfei Zhou$^{3}$\thanks{Corresponding author: \textit{Tianfei Zhou}.}   \\
	\small{$^1$ Northwestern Polytechnical University} \hspace{0pt}
	\small{$^2$ Beijing Institute of Technology} \hspace{0pt}
	\small{$^3$ Computer Vision Lab, ETH Zurich} \hspace{0pt} \\
	\small\url{https://github.com/zllxot/CORE}
}

\maketitle
\ificcvfinal\thispagestyle{empty}\fi

\begin{abstract}
	This paper presents \core, a conceptually simple, effective and communication-efficient model  for multi-agent cooperative perception.  It addresses the task from a novel perspective of \textbf{cooperative reconstruction}, based on two key insights: 1) cooperating agents together provide a more holistic observation of the environment, and 2) the holistic observation can serve as valuable supervision to explicitly guide the model  learning how to reconstruct the ideal observation based on collaboration.  \core instantiates the idea with three major components: a compressor for each agent to create more compact feature representation for  efficient broadcasting, a lightweight attentive collaboration component for cross-agent message aggregation, and a reconstruction module to  reconstruct the observation based on aggregated feature representations. This learning-to-reconstruct idea is task-agnostic, and offers clear and reasonable supervision to inspire more effective collaboration, eventually  promoting perception tasks. We validate \core on OPV2V, a large-scale multi-agent percetion dataset, in  two tasks, \ie, 3D object detection and semantic segmentation. Results demonstrate that \core achieves state-of-the-art performance, and is more communication-efficient.
\end{abstract}

\section{Introduction}

Perception -- identifying and interpreting sensory information -- is a crucial ability for intelligent agents to sense the surrounding environment. Thanks to continued advances in deep learning,  individual perception has demonstrated remarkable achievements in a number of tasks, \eg,   detection$_{\!}$ \cite{ren2015faster,zhou2018voxelnet,li2023human,meng2021towards},  segmentation$_{\!}$ \cite{long2015fully,wang2021exploring,zhou2022rethinking} and tracking$_{\!}$ \cite{wu2013online,zhou2022survey}.$_{\!}$ Though being  promising, it tends to suffer from issues (\eg, occlusion) stemmed from limited line-of-sight visibility of individual agents and is challenged by safety concerns. A more compelling paradigm  is \emph{cooperative perception}, \ie, a collection of agents to behave as a group by exchanging information with each other so as to use their combined sensory experiences to perceive. Along this direction,  recent efforts have been made to deliver datasets$_{\!}$ \cite{yu2022dair,xu2022opv2v,Li_2021_RAL,cui2022coopernaut} and cooperative solutions$_{\!}$ \cite{liu2020who2com,liu2020when2com,arnold2020cooperative,cui2022coopernaut,xu2022cobevt,su2022uncertainty,hu2022where2comm,chen2019f,wang2020v2vnet,xu2022v2x,xu2022opv2v}.

\begin{figure}[t]
	\begin{center}
		\includegraphics[width=\linewidth]{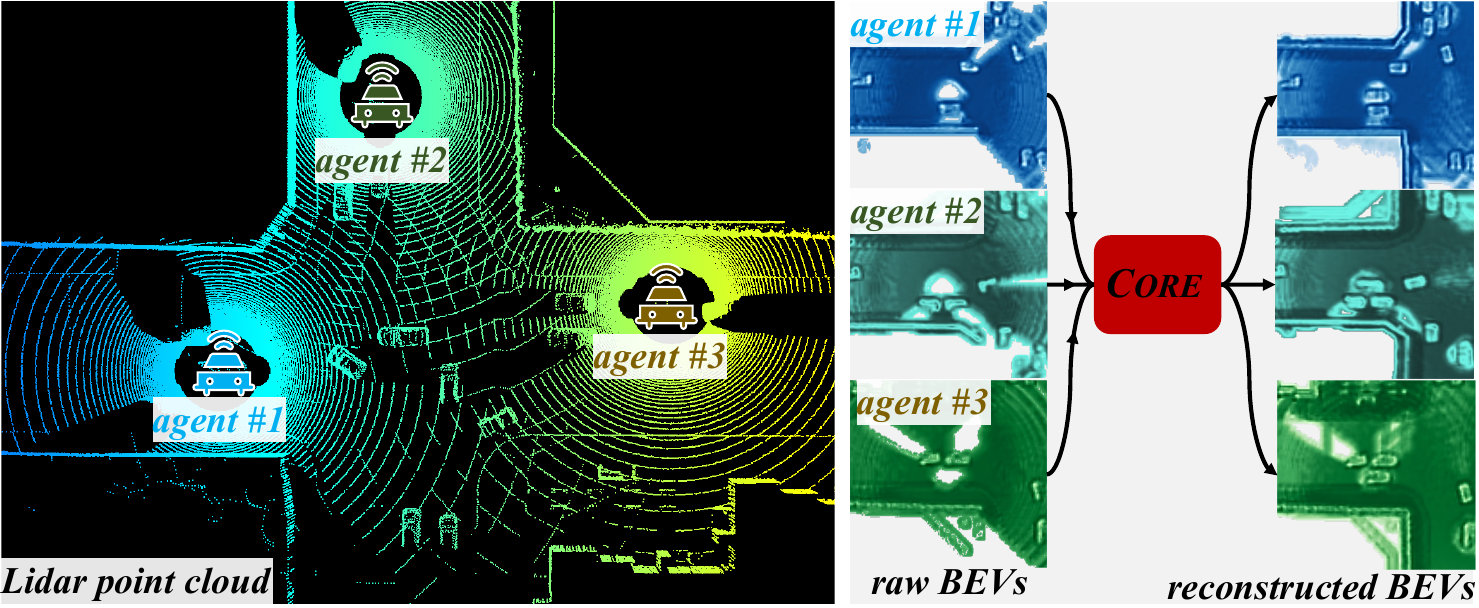}
	\end{center}
	\vspace{-15pt}
	\caption{\core addresses multi-agent perception from a novel perspective of cooperative reconstruction. Take the Lidar-based autonomous driving scenario as an example, individual vehicles are confined to limited sensing capabilities of onboard sensors,  yielding partial  observations of the whole scene (\ie, \emph{raw BEVs}). \core takes a straightforward but remarkably effective step to address this: in addition to task-aware learning (\eg, object detection, semantic segmentation), \core explicitly learns to reconstruct the complete scene (\ie, \emph{reconstructed BEVs}) from incomplete observations of cooperating agents. The reconstruction-aware learning objective serves as a more sensible goal to inspire  more effective cooperation of connected agents, ultimately boosting perception performance.}
	\label{fig:motivation}
\end{figure}

All these solutions hold a consensus promise that multiple agents together provide a holistic observation to the environment. But there is a practical challenge of performance-bandwidth trade-off to be addressed.  On this account, some studies explore  \emph{adaptive communication architectures} to reduce bandwidth requirements by dynamically determining for each agent, \eg, whom to communicate with \cite{liu2020who2com}, when to communicate \cite{liu2020when2com}, and what message to communicate \cite{hu2022where2comm}, while others focus more on the design of multi-agent \emph{collaborative strategies}, \eg, early collaboration \cite{chen2019cooper,arnold2020cooperative} to aggregate raw observations of cooperating agents, late collaboration \cite{liu2019fusioneye,volk2019environment} to combine  prediction results only, or the  prevailing intermediate collaboration \cite{chen2019f,wang2020v2vnet,xu2022cobevt,xu2022v2x,xu2022opv2v} to fuse  intermediate feature representations that are easy to compress. Although these approaches find applications in myriad domains, from autonomous driving \cite{Li_2021_RAL},  to automated warehouses \cite{sarkar2018scalable}, to search and rescue \cite{scherer2015autonomous}, a key question  remains unanswered:
\vspace{-16pt}
\begin{center}
	\emph{what should the ideal sensory state of each agent look like after information exchanging and aggregation?}
	\vspace{-4pt}
\end{center}
Prior approaches rely on task-specific objectives to learn how to communication or collaborative, which is sub-optimal and potentially degrades generalization capabilities of models to a broader array of perception tasks.

This paper advocates an approach from a novel perspective of cooperative reconstruction (see Fig.~\ref{fig:motivation}). Our main insight is that,  if multiple agents together indeed offer a more complete observation of scene, by absorbing others' information, an agent will be able to \textit{reconstruct} missing parts in its partial raw observation. By learning to reconstruct, the model is urged  to learn more effective task-agnostic feature representations, and
can offer a clearer explanation to the ideal cooperative state (\ie, features) of the agent, that is, from which we are able to reconstruct its complete  observation. Moreover, this learning-to-reconstruct idea naturally links to recent advancements in masked data modeling \cite{he2022masked,feichtenhofermasked,pang2022masked}, and  enables our model to recover complete observation from even more corrupted inputs obtained, \eg, by masking some proportion of raw observations. With this capability, agents will be allowed to exchange spatially sparse features to reduce transmission overhead during inference.

The idea is realized with our \core framework. Given a collection of agents and their 2D bird’s eye view (BEV) maps, \core performs cooperative reconstruction with three key modules: a \textit{{compression}} module, a \textit{{collaboration}} module and a \textit{{reconstruction}} module. 1) The {\textit{compression}} module computes a compressed feature representation of each BEV for efficient transmission. Unlike most prior works \cite{xu2022cobevt,li2021learning,xu2022v2x} that only consider channel-wise compression, the module imposes more significant compression by masking (sub-sampling) features along the spatial dimension. 2) The {\textit{collaboration}} module is a lightweight attention module that encourages knowledge exchanging across cooperating agents to enhance feature representations of each individual agent.  3) The {\textit{reconstruction}} module takes a decoder structure to recover complete scene observation from the enhanced feature representation. It is learned in a supervised rather than self-supervised  \cite{he2022masked,pang2022masked,feichtenhofermasked,li2022multi} manner, because we can easily obtain the supervision by aggregating raw sensor data of all agents. Note that the communication-costly raw data fusion only works in the training phase, and the entire reconstruction module will be discarded during inference. The reconstruction objective provides an explicit guidance to multi-agent collaboration, leading to ideal cooperative representations that will benefit various perception tasks. For task-specific decoding (\eg, detection or segmentation), \core applies conventional decoders in parallel with the reconstruction decoder.

To verify \core, we conduct experiments in the autonomous driving scenarios  on  the OPV2V~\cite{xu2022opv2v} dataset. Two popular  tasks, \ie, 3D object detection and semantic segmentation, are studied. Results show that \core is generalizable across tasks, and delivers better trade-off between perception accuracy and communication bandwidth.

\begin{figure*}[t]
	\begin{center}
		\includegraphics[width=\linewidth]{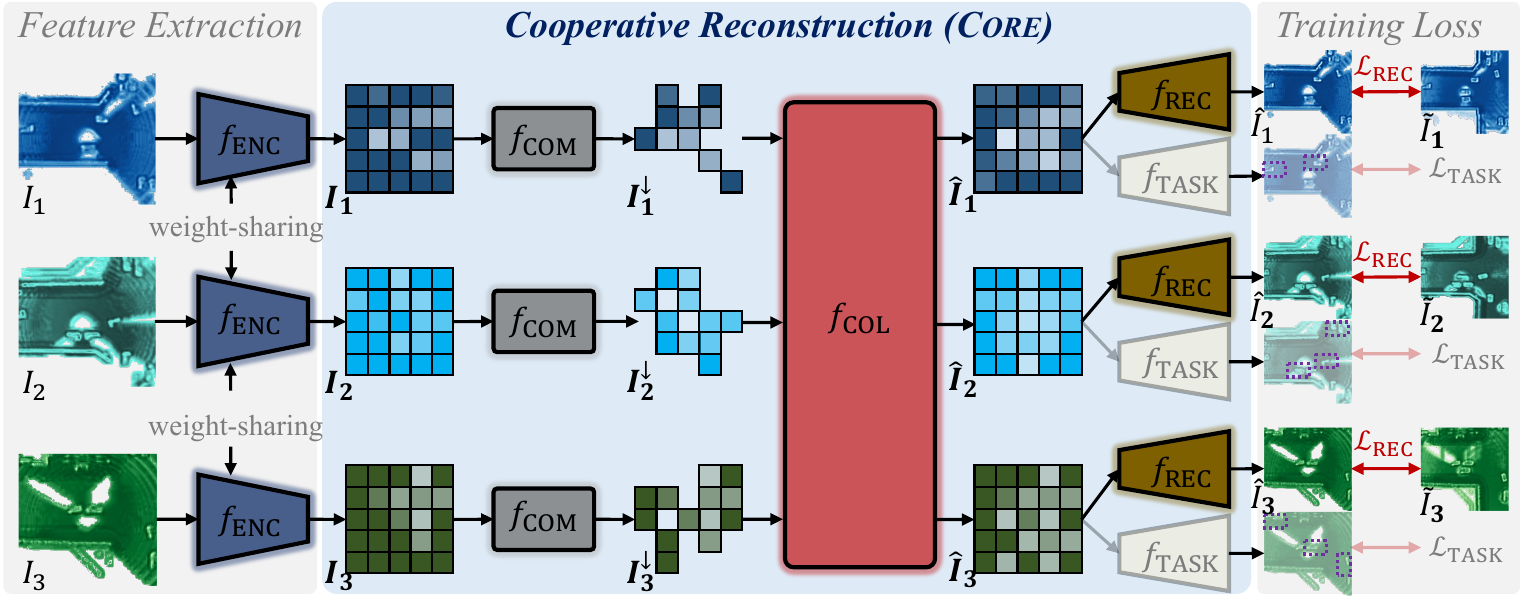}
	\end{center}
	\vspace{-17pt}
	\caption{\textbf{Illustration of \core} in  a three-agent cooperating scenario. Given raw BEV representations (\ie, $I_1$, $I_2$, $I_3$) of individual agents, \core achieves cooperative perception via several components: a shared $f_\text{ENC}$ for feature extraction, a compressor $f_\text{COM}$ for spatial- and channel-wise feature compression, a lightweight attentive collaborator $f_\text{COL}$ for information aggregation, a reconstruction decoder $f_\text{REC}$ to regress an ideal, complete BEV (\ie, $\hat{I}_1$, $\hat{I}_2$, $\hat{I}_3$), and a task-specific decoder $f_\text{TASK}$ for, \eg, object detection. \core is trained jointly by the reconstruction $\mathcal{L}_\text{REC}$ and task-specific $\mathcal{L}_\text{TASK}$ losses.}
		\vspace{-6pt}
	\label{fig:framework}
\end{figure*}

\section{Related Work}
\noindent\textbf{Cooperative Perception}  enables multiple agents to perceive surrounding environment collaboratively by sharing their observations and knowledge, which offers a great potential for improving individual's safety, resilience and adaptability \cite{lei2022latency,renrobust,han2023collaborative,cui2022coopernaut,li2021learning}. Recent years have witnessed many related systems developed to support a broad range of real-world applications, \eg,  vehicle-to-vehicle (V2V)-communication-aided autonomous driving \cite{xu2022v2x,xu2022opv2v,wang2020v2vnet,chen2019f,chen2019cooper,cui2022coopernaut},  multirobot warehouse automation system \cite{li2020mechanism,zaccaria2021multi} and multiple unmanned aerial vehicles (UAVs) for search and rescue \cite{scherer2015autonomous}. As single-agent perception tasks, the progress is largely driven by high-quality datasets, \eg,  OPV2V \cite{xu2022opv2v}, DAIR-V2X \cite{yu2022dair}, V2X-Sim 1.0 \cite{li2021learning} and 2.0 \cite{Li_2021_RAL}.

Based on these datasets, many effective cooperative perception methods are developed.  One straightforward solution, widely studed in early efforts \cite{chen2019cooper,arnold2020cooperative},  is to directly transmit and fuse raw sensor data among cooperating agents. Despite promising performance improvements, the expensive communication bandwidth required for transmitting high-dimensional data limits their deployment in practical situations. Recently, a new family of solutions~\cite{chen2019f,wang2020v2vnet,liu2020who2com,liu2020when2com,li2021learning,xu2022opv2v,hu2022where2comm,xu2022cobevt}, \ie, intermediate fusion-based methods, which broadcasts the compact intermediate representation, has been extensively studied as they are able to deliver a better trade-off between perception performance and network bandwidth. A mainstream branch~\cite{liu2020who2com,liu2020when2com,xu2022opv2v,hu2022where2comm,xu2022cobevt} of these methods is to employ attention mechanisms to obtain fusion weights, while other methods~\cite{wang2020v2vnet,li2021learning,zhou2022multi} model the relative relationships between different agents using graph neural networks.
Apart from performance, the robustness of system is also critical. Some methods~\cite{lei2022latency,vadivelu2021learning,renrobust} address problems arising from the communication process, for example, latency~\cite{lei2022latency,xu2022v2x}, localization error~\cite{vadivelu2021learning,yuan2022keypoints}, and communication interruption~\cite{renrobust}. Some other approaches~\cite{li2022multi,chen2022model} investigates self-supervised learning mechanisms to improve the generalization ability of collaboration models.

\core is closely relevant with DiscoNet~\cite{li2021learning} in terms that both methods rely on early collaboration to yield holistic-view inputs and leverage them as valuable guidance of network learning. However, \core addresses the problem based on a  learning-to-reconstruction formulation rather than teacher-student knowledge distillation in \cite{li2021learning}. In addition to yielding a simpler and more elegant model design, our formulation has two other  advantages: \textit{first}, by optimizing a reconstruction-aware objective, \core achieves better generalizibility as well as improved perception performance, \eg, it outperforms \cite{li2021learning} significantly by about $3\%$ on the task of BEV semantic segmentation (see Table~\ref{table:seg}); \textit{second}, the reconstruction idea allows \core to further masking out spatial features to be transimitted, thereby achieving more efficient communication against \cite{li2021learning}.

\section{Methodology}

The framework of \core is illustrated in Fig.~\ref{fig:framework}. We assume a system of $N$ agents simultaneously perceiving the environment. Each agent is equipped with an onboard sensor (\eg, RGB or thermal cameras, Lidar) to observe the environment and obtain its local measurement in form of modalities like RGB images or 3D point clouds. As conventions \cite{li2021learning,xu2022cobevt,hu2022where2comm},  \core explores perception in BEV space, where agents transform their individual perception information into BEV. Denote  ${I}_i\!\in\!\mathbb{R}^{h \times w \times c}$ as the BEV representation of the $i$-th agent with spatial size $h\!\times\!w$ and channel number $c$.  The agent encodes  ${I}_i$ into an intermediate feature representation by $\bm{I}_i\!=\! f_\text{ENC}({I}_i)\!\in\!\mathbb{R}^{H \times W \times C}$. Here, $f_\text{ENC}$ is a feature encoder shared by all agents, and $H$, $W$, $C$ denote feature height, width, channel, respectively. 

\core compresses the intermediate feature $\bm{I}_i$ of each agent independently to a compact representation $\bm{I}^{\downarrow}_i\!=\!f_\text{COM}(\bm{I}_i)$ using a shared compressor $f_\text{COM}$ (\S\ref{sec:com}) and broadcast it along with its pose to other agents. Subsequently, each agent $i$ aggregates the received features to enhance its representation $\hat{\bm{I}}_i\!=\!f_\text{COL}(\bm{I}_i^\downarrow, \{\bm{I}_j^\downarrow\}_{j\in[N]\setminus\{i\}})\!\in\!\mathbb{R}^{H\!\times\!W\!\times\!C}$ based on an attentive collaboration component (\S\ref{sec:col}). Unlike existing methods that learn $\hat{\bm{I}}_i$ in a task-specific manner, \core provides additional guidance to learn the ideal state of $\hat{\bm{I}}_i$ from which the  BEV representation can be reconstructed, denoted as $\hat{I}_i\!=\!f_\text{REC}(\hat{\bm{I}}_i)\!\in\!\mathbb{R}^{h \times w \times c}$ (\S\ref{sec:rec}). Here $f_\text{REC}$ is a reconstruction decoder. Next, we elaborate on  details of the key modules.

\subsection{Feature Compression and Sharing} \label{sec:com}
The goal of feature compression is to reduce the communication bandwidth as much as possible while maintaining perception performance. Previous approaches only compress features (\eg, $\bm{I}_i$ for agent $i$) along the channel dimension by, \eg, $1\times1$ convolutional autoencoder \cite{wang2020v2vnet,li2021learning,xu2022v2x}. 
However, we argue that simply compressing the channel dimension does not lead to the maximal saving of the bandwidth. To drive more effective reconstruction, \core additionally performs spatial-wise sampling to features, which can further alleviate the transmission burden. Concretely, for channel-wise compression, we follow  \cite{xu2022v2x} to use a series of $1\times1$ convolutional layers to progressively compress $\bm{I}_i$, and the compressed feature is with the shape $(H, W, C')$ and $C'\!\ll\!C$. In our experiments, we follow prior studies \cite{li2021learning,xu2022opv2v,xu2022v2x} to set $C'=16$. For spatial-wise sub-sampling, 
a two-step scheme is taken: first, we sum up the channel-compressed feature map along the channel dimension  to obtain a 2D activation map $\bm{O}_i\!\in\!\mathbb{R}^{H \times W}$ and select the top $K\%$ of the $HW$ feature points with the highest activation values; second, we further sample $R\%$ of the $(HWK)\%$ points following a uniform distribution, leading to a set of $(HWKR)\text{\textpertenthousand}$ feature points at final. These features and their corresponding spatial coordinates in $\bm{I}_i$ are messages that will be transmitted to connected agents nearby. After other agents receiving the feature points, they will reorganize them into an all-zero-initialized feature map $\bm{I}_i^\downarrow\!\in\!\mathbb{R}^{H\times W \times C'}$, and then project them back to $(H, W, C)$ using $1 \times 1$ convolutions.

\subsection{Attentive Collaboration} \label{sec:col}

Multi-agent collaborative learning focuses on updating each agent's feature map by aggregating informative messages from its partners. Without loss of generality, we assume the $i$-th agent as the ego agent, and we have its local feature map $\bm{I}_i^\downarrow$ as well as the features received from connected agents $\{\bm{I}_j^\downarrow\}_{j=1}^J$. Here $J$ denotes the  number of agents connected to $i$. To ensure the sender and the receiver are properly aligned before collaboration, we geometrically warp the sender's message $\bm{I}_j^\downarrow$ onto the receiver's coordinate to derive $\bm{I}_{j \rightarrow i}^\downarrow \!=\! {\Gamma}_\xi(\bm{I}_j^\downarrow)$. Here ${\Gamma}_\xi$ is a differential transformation and sampling operator \cite{xu2022v2x}.

\begin{figure}[t]
	\begin{center}
		\includegraphics[width=\linewidth]{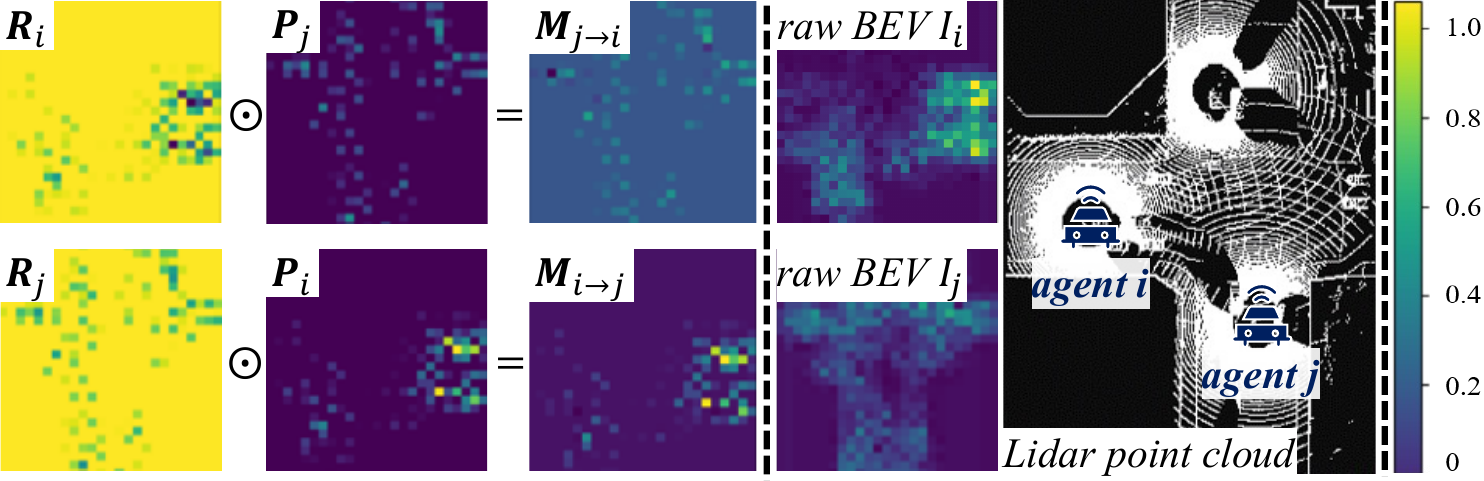}
	\end{center}
	\vspace{-15pt}
	\caption{\textbf{Visualization} of computations of $\bm{M}_{j \rightarrow i}$ and $\bm{M}_{i \rightarrow j}$ in Eq.~\ref{eq:att}. Point cloud of the holistic scene and raw BEVs of individual agents are shown for reference.}
		\vspace{-6pt}
	\label{fig:pr}
\end{figure}

For multi-agent collaboration, we are inspired by \cite{hu2022where2comm} and  devise a lightweight attention-aware collaboration scheme.  Concretely, for each agent $i$, we transform its feature $\bm{I}_i^\downarrow$ using a series of $1\!\times\!1$ convolutions followed by a softmax function to generate a confidence map $\bm{P}_i \!\in\! [0, 1]^{H\times W}$. Each element in $\bm{P}_i$ represents the activation degree at the corresponding spatial location, \ie, higher values are more objective-sensitive. $\bm{P}_i$ conveys \emph{what information agent $i$ can offer to others}. Moreover, we  compute another confidence map  $\bm{R}_i\in[0, 1]^{H\times W}$ by $\bm{R}_i = 1-\bm{P}_i$. In contrast to $\bm{P}_i$, the spatial locations with higher values in $\bm{R}_i$ indicate there is potentially  information missing caused by occlusions or limited visibility, hence, it reflects \textit{what information agent $i$ needs the most}. Given the confidence maps, for ego agent $i$ and its partner agent $j$, we compute an attention map as follows (see Fig.~\ref{fig:pr}):
\begin{equation}\label{eq:att}
	\bm{M}_{j \rightarrow i} = \bm{R}_i\odot \bm{P}_j~~~\in [0, 1]^{H\times W},
\end{equation}
where $\odot$ is the Hadamard product. Here the attention map $\bm{M}_{j \rightarrow i}$ highlights the positions that agent $i$ needs information while agent $j$ can meet $i$'s requirements. The attention map allows us to perform more accurate and adaptive feature aggregation.

Furthermore, we update the feature $\bm{I}_{i}^\downarrow$ of ego agent $i$ based on the calibrated feature $\bm{I}_{j\rightarrow i}^\downarrow$ from agent $j$ and attention mask $\bm{M}_{j \rightarrow i}$ as follows:
\begin{equation}\label{eq:attention}
	\!\!\hat{\bm{I}}_{j\rightarrow i} \!=\! \mathrm{DConv}_{l\times l}(\bm{A})\odot \bm{V}\odot \bm{M}_{j \rightarrow i} \!+\! \bm{I}_{i}^\downarrow ~~~\in \mathbb{R}^{H\times W\times C}, \!\!
\end{equation}
where $\bm{A}$ and $\bm{V}$ are computed as:
\begin{align}
	\bm{A} &= \bm{W}_1[\bm{I}_{i}^\downarrow, \bm{I}_{j\rightarrow i}^\downarrow]~~\in \mathbb{R}^{H\times W\times 2C}, \\
	\bm{V} &= \bm{W}_2\bm{I}_{j\rightarrow i}^\downarrow ~~\in \mathbb{R}^{H\times W\times C}.
\end{align}

\begin{figure}[t]
	\begin{center}
		\includegraphics[width=\linewidth]{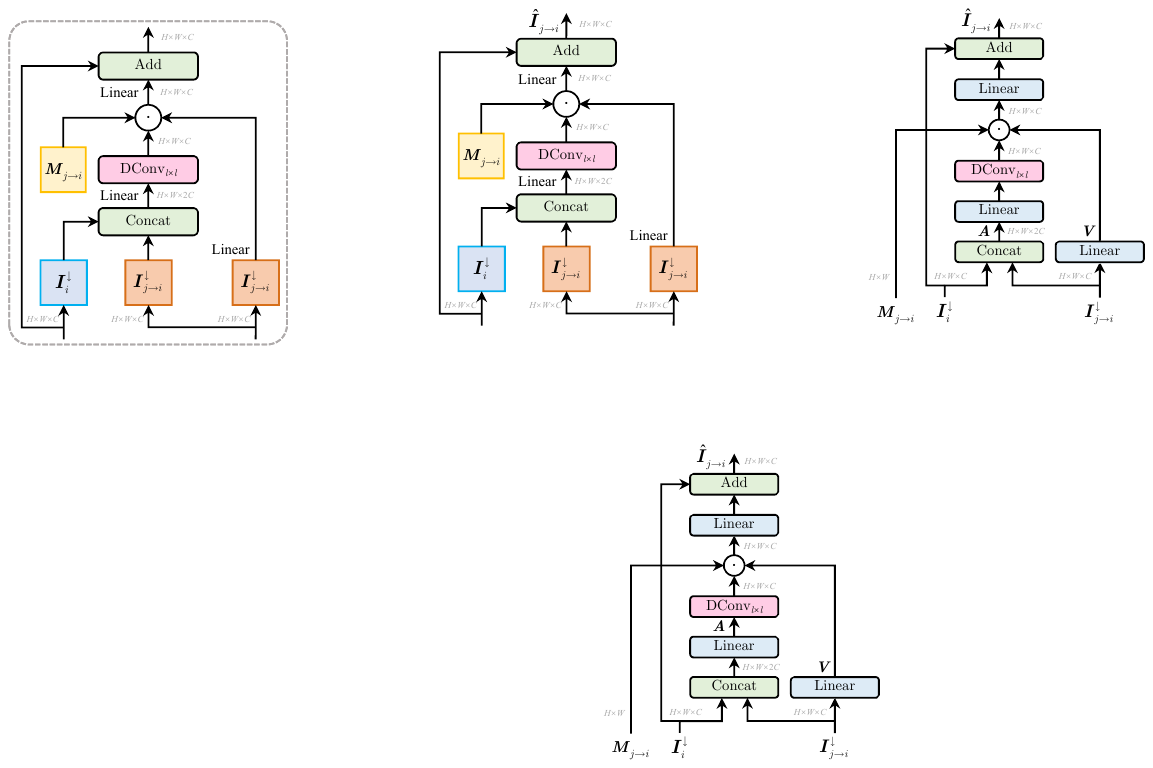}
	\end{center}
	\vspace{-15pt}
	\caption{\textbf{Structure} of attentive collaboration (Eq.~\ref{eq:attention}).}
		\vspace{-6pt}
	\label{fig:attention}	
\end{figure}

Fig.~\ref{fig:attention} illustrates the structure of the attentive collaboration scheme.
Here  $\hat{\bm{I}}_{j\rightarrow i}$ is the updated message of agent $i$ based on the information from agent $j$, $\mathrm{DConv}_{l\times l}$ indicates depthwise convolution with  kernel size $l\times l$ that generates an output with half the number of input channels. It enhances the receptive field of  features while reducing the complexity and computational cost. `$[\cdot, \cdot]$' is a channel-wise tensor concatenation operator, combining the message from different agents. $\bm{W}_1$ and $\bm{W}_2$ are learnable linear weights that allow cross-channel information interaction. The first term in Eq.~\ref{eq:attention} computes the attention weight from agent $j$ to $i$, where the attention map $\bm{M}_{j \rightarrow i}$ makes it more targeted. A residual layer is introduced to avoid information loss in $\bm{I}_i^\downarrow$.

For ego agent $i$, Eq.~\ref{eq:attention} is individually applied to each of its connected agent. Its final feature  is  the average of  updated features from all agents, \ie, $\hat{\bm{I}}_i=\texttt{avg}_j(\{\hat{\bm{I}}_{j\rightarrow i}\})$.

In this way, our attentive collaboration module can effectively capture spatial and channel-wise dependencies among agents and learns to selectively attend to the most informative messages from neighboring agents. By  exchanging messages across all connected agents, the proposed scheme enables each agent to refine its own representation by taking into account the context of the whole group and eventually improve the overall performance of the group on the target task.

\begin{figure}[t]
	\begin{center}
		\includegraphics[width=\linewidth]{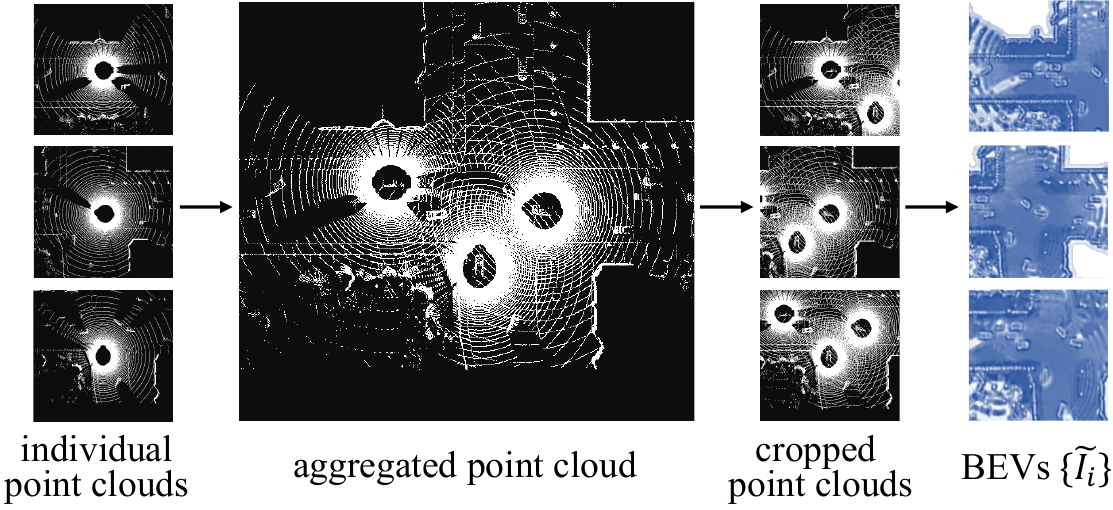}
	\end{center}
	\vspace{-15pt}
	\caption{Illustration of the creation of ideal BEVs $\{\tilde{I}_i\}$ based on point cloud aggregation. See \S\ref{sec:rec} for details.}
		\vspace{-8pt}
	\label{fig:agg}
\end{figure}

\subsection{BEV Reconstruction}\label{sec:rec}

In \S\ref{sec:com} and \S\ref{sec:col}, we reduce the bandwidth consumption by compressing the intermediate representation along the channel and spatial dimensions, and achieve efficient information interaction across agents through attentive collaboration. However, the detailed information contained in intermediate representations is inevitably lost due to compression. In addition, supervising collaborative learning solely through task-specifc labels can lead to a high dependence on downstream tasks, which hurts the generalization ability of the model. Our BEV reconstruction scheme alleviates these issues by learning a deep convolutional decoder  to reconstruct the multi-view BEV representations from the interacted messages.

To obtain proper supervisory signal, we aggregate raw point clouds of individual agents to yield a holistic-view point cloud and convert into BEV representations that will serve as  prediction targets of the reconstruction decoder. A conceptual illustration is presented in Fig.~\ref{fig:agg}.
To be more specific, let $\bm{S}_i$ be the raw 3D point clouds collected from the ego agent $i$. We first  project point clouds from all neighbor agents, $\{\bm{S}_j\}_{j=1}^J$, to the ego's coordinate system: $\{\bm{S}_{j\rightarrow i}\}_{j=1}^J = \Gamma_\xi(\{\bm{S}_j\}_{j=1}^J)$. Next, we aggregate each individual point cloud to produce a multi-view 3D scene: $\tilde{\bm{S}} = f_{\text{Sta}}(\{\bm{S}_{j\rightarrow i}\}_{j=1}^J, \bm{S}_i)$, where $f_{\text{Sta}}(\cdot, \cdot)$ is a stacking operator. Now, all agents are situated within the same coordinate system yet occupy distinct positions. Subsequently, we project back to each local coordinate and crop it based on perceptual range: $\tilde{\bm{S}}_i = \Gamma^{-1}_{\xi}(\tilde{\bm{S}})$. 
Finally, we convert $\tilde{\bm{S}}_i$ into its corresponding BEV representations $\tilde{{I}_i}\in \mathbb{R}^{h\times w\times c}$.
Here, $\tilde{{I}_i}$ has the same dimension as the original BEV feature maps $I_i$, but offers  a much broader viewpoint.

For each agent $i$, the reconstruction decoder takes the updated feature $\hat{\bm{I}}_i$ as input and reconstructs the corresponding BEV feature as follows:
\begin{equation}
	\hat{I}_i\!=\!f_\text{REC}(\hat{\bm{I}}_i)\!\in\!\mathbb{R}^{h \times w \times c}.
\end{equation}

To encourage the reconstructed data to match the original input data, we train our reconstruction decoder using a reconstruction loss. Specifically, given the reconstructed BEV features $\hat{{I}_i}$ w.r.t its supervision signal $\tilde{{I}_i}$, the reconstruction loss is computed as the mean squared error (MSE) between them:
\begin{equation}
	\mathcal{L}_\text{REC} = \sum_{x=1}^{h}\sum_{y=1}^{w}||\hat{{I}_i}(x, y)-\tilde{{I}_i}(x, y)||^2_2.
	\label{eq:rec}
\end{equation}

By incorporating feature reconstruction into our pipeline, we are able to effectively leverage the benefits of collaboration without sacrificing the independence and flexibility of each individual task.

\subsection{Detailed Network Architecture}

\noindent\textbf{Network Architecture.} The entire network is comprised of three major components: 
\begin{itemize}[leftmargin=*]
	\setlength{\itemsep}{0pt}
	\setlength{\parsep}{-2pt}
	\setlength{\parskip}{-0pt}
	\setlength{\leftmargin}{-10pt}
	\vspace{-4pt}
	
	\item \textbf{Feature Extractor $f_\text{ENC}$.} The feature extractor $f_\text{ENC}$ takes the BEV representation as input and encodes it into intermediate features. Following~\cite{xu2022opv2v}, we implement it through three 2D convolutional layers with kernel size of $3\!\times\!3$, and a batch normalization layer and a ReLU activation layer are employed after each layer.
	
	\item \textbf{Reconstruction Decoder $f_\text{REC}$.}
	The reconstruction decoder $f_\text{REC}$ is used for generating high-quality BEV features. It  consists of three blocks, each of which is composed of a $2\!\times\!2$ transposed convolutional layer and a $3\!\times\!3$ normal convolutional layer. Each layer is followed by a batch normalization layer and a ReLU activation layer.
	
	\item \textbf{Task-specific Decoder $f_\text{TASK}$.} We study two tasks: 3D object detection and semantic segmentation. For \textit{detection}, the decoder comprises three deconvolutional layers followed by two sibling branches, \ie, class prediction and bounding box regression. 
	Following \cite{xu2022opv2v}, the class prediction branch produces the confidence score for each anchor box, distinguishing between being an object or background, while the regression output represents the predefined anchor boxes' attributes, including their location, size, and orientation.
	For \textit{segmentation}, the decoder consists of three deconvolutional layers for upsampling and a $3\!\times\!3$ convolutional layer for generating the final semantic segmentation map.
\end{itemize}

\noindent\textbf{Loss Function.}
\core is optimized by minimizing the following  loss function:
\begin{equation}\label{eq:loss}
	\mathcal{L} = \mathcal{L}_\text{TASK} + \lambda\mathcal{L}_\text{REC},
\end{equation}
where the first term $\mathcal{L}_\text{TASK}$ is a task-specific loss. For detection, we follow  conventions~\cite{li2021learning,xu2022opv2v} to use focal loss as the classification loss and smooth $l_1$ loss for bounding box regression. For segmentation, we use cross-entropy loss as~\cite{xu2022cobevt}. The second one is our reconstruction loss $\mathcal{L}_\text{REC}$ (Eq.~\ref{eq:rec}) for supervising the output of reconstruction decoder. The coefficient $\lambda$ balances the two loss terms.

\section{Experiment}

For comprehensive evaluation, we validate \core on two popular cooperative perception tasks, \ie, 3D object detection and BEV semantic segmentation.
\subsection{Experimental Setup}
\noindent\textbf{Datasets.} Our experiments are conducted on OPV2V~\cite{xu2022opv2v}, which is a large-scale public dataset collected together by the co-simulating framework OpenCDA~\cite{xu2021opencda} and the CARLA~\cite{dosovitskiy2017carla} simulator for vehicle-to-vehicle collaborative perception. In total, the dataset contains $11,\!464$ frames of LiDAR point clouds and RGB images.
OPV2V can be further split into two subsets: a default CARLA towns and a Culver City digital town. The default CARLA towns subset includes $10,\!914$ frames ($6,\!764/1,\!980/2,\!170$ for \texttt{train}/\texttt{val}/\texttt{test} splits). This subset provides a diverse range of scenarios with different levels of complexity, which can be used to train and evaluate collaborative perception models. The Culver City subset includes $550$ frames and is designed to evaluate the generalization ability of the models. This subset represents a realistic imitation of a real-world city, with a variety of objects and structures that can challenge the perception capabilities of the models.

\noindent\textbf{Training.} During the training phase, a set of agents that can establish communication with each other is randomly selected from the scenario, with each agent is assigned a communication range of $70$ m. We limit the range of point clouds to $[-140.8, 140.8]\times[-40, 40]\times[-3, 1]$ along the $x$, $y$, and $z$-axis for the 3D object detection task, and to $[-51.2, 51.2]\times[-51.2, 51.2]\times[-3, 1]$ for the BEV semantic segmentation task. 
As the voxel resolution is set to $0.4$ m, we can get the BEV map with a resolution of $200\times704\times128$ and $256\times256\times128$, respectively. We apply several data augmentation techniques to the training data, including random flipping, scaling in range of $[0.95, 1.05]$, and rotation from $[-45^{\circ}, 45^{\circ}]$. We train our \core using Adam~\cite{kingma2014adam} optimizer with a learning rate of $0.002$ and a batch size of $1$. We also employ early stopping based on the validation loss to prevent overfitting. The hyperparameters $\lambda$, $R$ and $K$ are empirically set to $1$, $90$ and $90$, respectively.

\noindent\textbf{Inference.} At the inference stage, we adopt a fixed group of agents to evaluate the model for fair comparison. We apply a confidence score threshold of $0.25$ and non-maximum suppression (NMS) with an IoU threshold of $0.15$ to filter out overlapping detections.

\noindent\textbf{Baselines.} For comparison, we build two baseline models. \textit{No Collaboration} refers to a single-agent perception system using only individual sensor data without any information from other agents. 
\textit{Early Collaboration} directly aggregates raw sensor data from multiple agents at an early stage of the perception pipeline.
\textit{Late Collaboration} collects the prediction results of multiple agents and combines them to deliver the final results using NMS. In addition, \core is compared with  existing state-of-the-art algorithms, including \textit{Cooper}\!~\cite{chen2019cooper}, \textit{F-Cooper}\!~\cite{chen2019f}, \textit{V2VNet}\!~\cite{wang2020v2vnet}, \textit{DiscoNet}\!~\cite{li2021learning}, \textit{AttFuse}\!~\cite{xu2022opv2v}, and \textit{CoBEVT}\!~\cite{xu2022cobevt}.

\noindent\textbf{Evaluation Metrics.} For  3D object detection, we adopt standard evaluation metrics as~\cite{li2021learning,xu2022opv2v}: average precision (AP) at intersection-over-union (IoU) thresholds of $0.5$ and $0.7$. For  BEV semantic segmentation, we follow~\cite{xu2022cobevt} to report IoU scores for categories of vehicle, road, and lane.

\noindent\textbf{Reproducibility.} We implemented \core using PyTorch and trained it on an NVIDIA RTX 3090Ti GPU with a 24GB  RAM. Testing is conducted on the same machine.

\newcommand{\reshl}[2]{
	\textbf{#1} \fontsize{7.5pt}{1em}\selectfont\color{mygreen}{$\uparrow$ \textbf{#2}}
}

\begin{table}
	\centering
	\small
	\resizebox{0.49\textwidth}{!}{
		\setlength\tabcolsep{1pt}
		\renewcommand\arraystretch{1.05}
		\begin{tabular}{r||c|cc|cc}
			\rowcolor{mygray}
			\thickhline
			\multirow{2}{*}{} & \multirow{2}{*}{} & \multicolumn{2}{c|}{{Default}} & \multicolumn{2}{c}{{Culver}}\\
			\rowcolor{mygray}
			\multirow{-2}{*}{{Method}} & \multirow{-2}{*}{{Backbone}} & AP@$0.5$ & AP@$0.7$ & AP@$0.5$ & AP@$0.7$\\ \hline\hline
			No Collaboration    & \multirow{7}{*}{VoxelNet~$_{\!}$\cite{zhou2018voxelnet}} & 68.8 & 52.6 & 60.5 & 43.1 \\
			Early Collaboration &  & 89.9 & 85.8 & 87.7 & 78.4 \\
			Late Collaboration &  & 80.1 & 73.8 & 72.2 & 58.8  \\
			Cooper\!~\cite{chen2019cooper} &  & 85.2 & 75.8 & 81.5 & 67.7 \\
			F-Cooper\!~\cite{chen2019f} &  & 87.6 & 78.7 & 84.9 & 72.3 \\
			AttFuse\!~\cite{xu2022opv2v} &  & \textbf{90.9} & 85.2 & 84.3 & 74.7 \\
			\core &  & \textbf{90.9} & \textbf{88.3} & \textbf{87.8} & \textbf{82.6} \\\hline
			No Collaboration    & \multirow{9}{*}{PointPillar~$_{\!}$\cite{lang2019pointpillars}} & 67.9 & 60.2 & 55.7 & 47.1 \\
			Early Collaboration &  & 89.3 & 83.5 & 86.1 & 75.9 \\
			Late Collaboration &  & 85.8 & 78.1 & 79.9 & 66.8  \\
			V2VNet\!~\cite{wang2020v2vnet}   &  & 89.7 & 82.2 & 86.8 & 73.3 \\
			Cooper\!~\cite{chen2019cooper}   &  & 89.1 & 80.0 & 82.9 & 69.6 \\
			F-Cooper\!~\cite{chen2019f}   &  & 88.7 & 79.1 & 84.5 & 72.9 \\
			AttFuse\!~\cite{xu2022opv2v} &  & 89.9 & 81.1 & 85.4 & 73.6 \\
			CoBEVT\!~\cite{xu2022cobevt} &  & \textbf{91.4} & \textbf{86.2} & 85.9 & 77.3 \\
			\core &  & 90.9 & 85.8 & \textbf{87.7} & \textbf{78.1} \\\hline
		\end{tabular}
	}
	\vspace{-5pt}
	\caption{\small\textbf{Quantitative results} for the task of 3D object detection  on OPV2V \cite{xu2022opv2v}.}
		\vspace{-8pt}
	\label{table:det}
\end{table}

\begin{table}
	\centering
	\small
	\resizebox{0.49\textwidth}{!}{
		\setlength\tabcolsep{4pt}
		\renewcommand\arraystretch{1.05}
		\begin{tabular}{r||c|ccc|c}
			\rowcolor{mygray}
			\thickhline
			{Method} & {Backbone} & {Vehicle} & {Road} & {Lane} & Avg.\\ \hline\hline
			No Collaboration & \multirow{8}{*}{VoxelNet~\cite{zhou2018voxelnet}} & 35.4 & 55.5 & 40.3 & 43.7 \\
			Early Collaboration &  & 53.5 & 60.6 & 44.1 & 52.7 \\
			Late Collaboration &  & 44.6 & 59.6 & 42.5 & 48.9 \\
			F-Cooper\!~\cite{chen2019f} &  & 53.9 & 61.0 & 46.8 & 53.9 \\
			DiscoNet\!~\cite{li2021learning} &  & 54.1 & 61.1 & 46.6 & 53.9 \\
			AttFuse\!~\cite{xu2022opv2v} &  & 54.6 & 61.3 & 47.1 & 54.3 \\
			V2VNet\!~\cite{wang2020v2vnet} &  & 58.4 & 60.7 & 45.9 & 55.0 \\
			\core &  & \textbf{60.8} & \textbf{62.0} & \textbf{47.7} & \textbf{56.8} \\ \hline
		\end{tabular}
	}
	\vspace{-5pt}
	\caption{\small\textbf{Quantitative results} for the task of BEV semantic segmentation on OPV2V \cite{xu2022opv2v}.}
	\vspace{-8pt}
	\label{table:seg}
\end{table}

\begin{figure*}[t]
	\begin{center}
		\includegraphics[width=\linewidth]{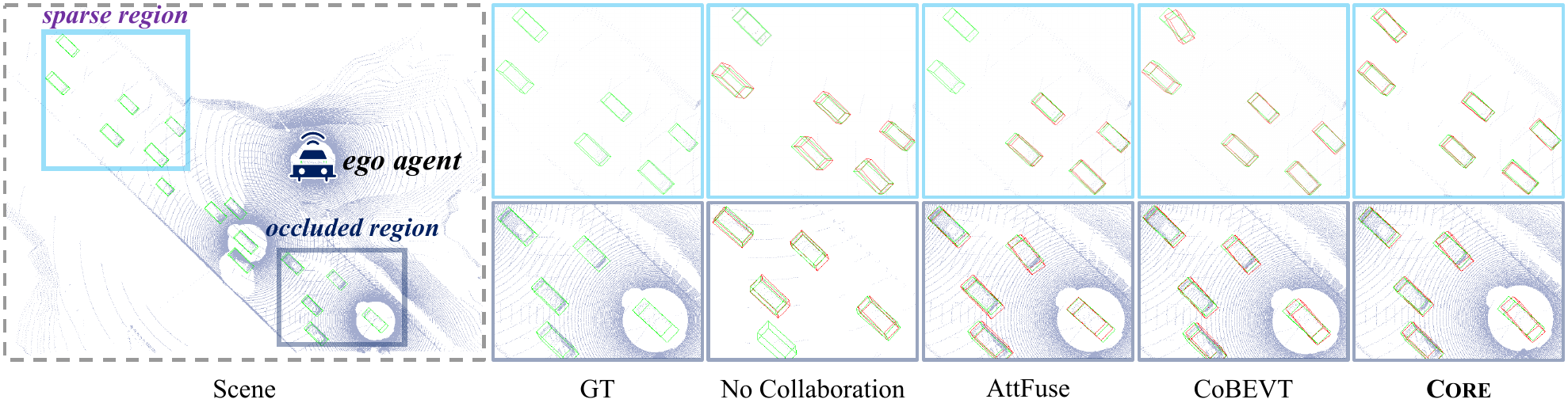}
	\end{center}
	\vspace{-15pt}
	\caption{\small\textbf{Qualitative object detection results} on OPV2V~\cite{xu2022opv2v}. \core is compared with \textit{No collaboration}, \textit{AttFuse} \cite{xu2022opv2v} and \textit{CoBEVT} \cite{xu2022cobevt} on two regions with sparse observations and occlusions. }
	\label{fig:visual}
\end{figure*}

\begin{figure*}[t]
	\begin{center}
		\includegraphics[width=\linewidth]{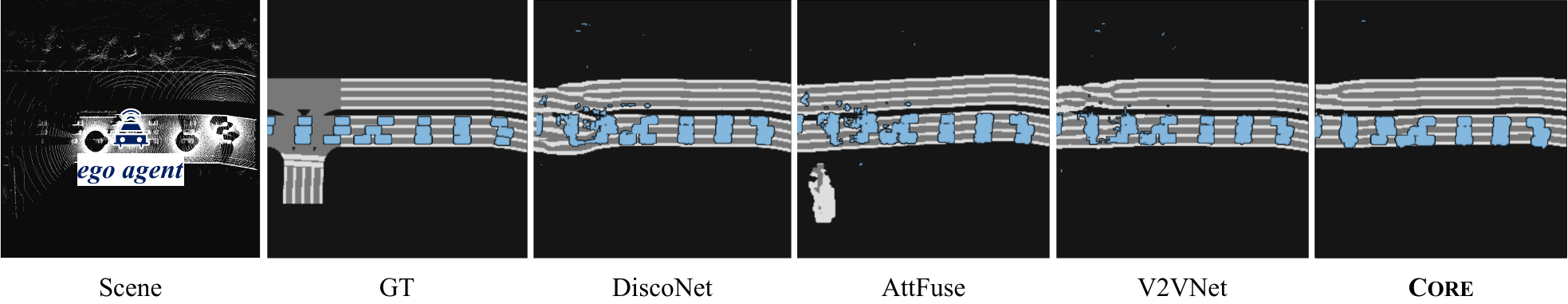}
	\end{center}
	\vspace{-15pt}
	\caption{\small\textbf{Qualitative segmentation results} on OPV2V~\cite{xu2022opv2v}. \core is compared with \textit{DiscoNet} \cite{li2021learning}, \textit{AttFuse} \cite{xu2022opv2v} and \textit{V2VNet} \cite{wang2020v2vnet}.}
	\vspace{-9pt}
	\label{fig:seg}
\end{figure*}

\subsection{Quantitative Result}
\noindent\textbf{3D Object Detection.} Table~\ref{table:det} reports the overall detection results of  \core against existing approaches in terms of AP@$0.5$ and AP@$0.7$ on  OPV2V \cite{xu2022opv2v}. For comprehensive evaluation, we evaluate methods using two different types of 3D perception backbones, \ie, a voxel-based method VoxelNet~\cite{zhou2018voxelnet} and a point-based method PointPillar~\cite{lang2019pointpillars}. Regarding VoxelNet-based methods, the table shows that \core yields state-of-the-art performance across all the metrics on both subsets and  surpasses the previous state-of-the-art method, \textit{AttFuse}~\cite{xu2022opv2v}, by a significant margin of, \eg, \textbf{3.5\%} and \textbf{7.9\%} in terms of AP@$0.5$ and $0.7$ on Culver City. Among methods based on PointPillar backbone, \core consistently outperforms \textit{AttFuse}~\cite{xu2022opv2v} across all the metrics. Compared with a more recent method \textit{CoBEVT}~\cite{xu2022cobevt}, \core outperforms it by solid margins of \textbf{1.8\%}/\textbf{0.8\%} in terms of AP@$0.5$/$0.7$ on Culver City.  Though \cite{xu2022cobevt} is better than \core on Default subset, the gap is indeed minor ($<0.5\%$).  In summary, these results suggest that \core is very robust to backbones, and outperforms current state-of-the-art methods in terms of AP@$0.5$ and AP@$0.7$, especially on the Culver City subset. 

\noindent\textbf{BEV Semantic Segmentation.} Table~\ref{table:seg} presents BEV semantic segmentation results. For fair comparison, we use the same backbone  (\ie, VoxelNet) and segmentation head for all methods. The table shows that our proposed method \core outperforms all other methods by a significant margin, achieving an average IoU score of $\textbf{56.8\%}$, clearly outperforming the second-best method \textit{V2VNet} \cite{wang2020v2vnet} with an average score of 55.0\%.   \core  consistently achieves the best performance for all three categories, demonstrating its ability to handle different segmentation targets.  Furthermore, our proposed method shows a significant improvement over the baselines with \textit{No Collaboration} and \textit{Late Collaboration}. It improves the \textit{No Collaboration} baseline by \textbf{13.1\%} and the \textit{Late Collaboration} baseline by \textbf{7.9\%}. These results demonstrate the effectiveness of \core in improving the performance of BEV semantic segmentation through collaborative learning. The segmentation results further confirms the perception superiority of \core, and further demonstrates its strong generalization ability.

\subsection{Qualitative Result}
We present qualitative results to visually illustrate the effectiveness of \core in handling challenging scenarios. As depicted in Fig.~\ref{fig:visual}, our algorithm, \core, can successfully detect multiple objects in cluttered, sparse point clouds, and occluded environments, and show better results in comparison with strong baselines like \textit{AttFuse} \cite{xu2022opv2v} and \textit{CoBEVT} \cite{xu2022cobevt}. To further validate the generalization capability of \core, we conducted experiments to evaluate the performance of \core on BEV semantic segmentation, as shown in Fig.~\ref{fig:seg}. We demonstrate that \core achieves high-quality segmentation results even in complex scenarios such as crowded and dynamic environments. These results are consistent with our quantitative evaluations, positioning it as a promising solution for various real-world applications.

\subsection{Ablation Study}
\noindent\textbf{Key Component Analysis.}  Table~\ref{table:key} shows the contribution of individual components in our \core framework for both detection and segmentation tasks. By adding the attentive collaboration (\ie, `Colla.') module alone, we observe an improvement in AP@$0.7$ and IoU scores, with a slight impact on AP@$0.5$ compared to the base case. The addition of the BEV reconstruction (\ie, `Recon.') module on top of the collaboration module leads to further improvement across all metrics, particularly in Culver City, where we see a significant increase of \textbf{2.8\%} in AP@$0.5$ and \textbf{1.6\%} in AP@$0.7$. These results highlight the effectiveness of the reconstruction module in enhancing the performance of our \core framework. Moreover, incorporating the reconstruction module also leads to a significant gain in segmentation performance, with a \textbf{1.5\%} increase in IoU compared to the baseline with collaboration. These findings demonstrate the importance of both collaboration and reconstruction modules in achieving state-of-the-art performance on multi-agent perception tasks.

\noindent\textbf{Spatial-wise Feature Compression.} Furthermore, we analyze the effect of feature compression in \core. Unlike most existing methods (\eg, \textit{DiscoNet} \cite{li2021learning}) only exploit channel-wise compression, \core additionally carries out spatial-wise feature compression. Table~\ref{table:compress} provides detection performance and transmitted data size under varied spatial compression ratio (we do not discuss channel-wise compression since it is same to most existing studies.) It is not surprised to see that performance progressively degrades as spatial compress ratio increases  from $1.0$ to $0.1$. But it is favorable that \core is able to deliver very promising performance even with a large compression ratio of $0.4$. Moreover, \core with a ratio of $0.6$ can achieve comparable performance as \textit{DiscoNet} \cite{li2021learning} with no compression. For a better trade-off between perception performance and communication efficiency, we set the spatial compression ratio to $0.8$ by default.

\begin{table}[t]
	\begin{centering}
		\small
		\resizebox{0.49\textwidth}{!}{
			\setlength\tabcolsep{6pt}
			\renewcommand\arraystretch{1.05}
			\begin{tabular}{cc||cc|cc}
				\rowcolor{mygray}
				\thickhline
				{Colla.} & Recon. & \multicolumn{2}{c|}{{Object Detection}} & {Segmantation} \\
				\rowcolor{mygray}
				(\S\ref{sec:col}) & (\S\ref{sec:rec}) & AP@$0.5$ & AP@$0.7$ & Vehicle\\
				\hline \hline
				&  & 90.9 / 84.3 & 85.2 / 74.7 & 54.6 \\
				\ding{51}	&  & 90.8 / 84.4 & 86.1 / 77.1 &  56.8 \\
				\ding{51}	& \ding{51} &  \textbf{91.1} / \textbf{87.2} & \textbf{86.9} / \textbf{78.7}  & \textbf{58.3} \\ \hline
			\end{tabular}
		}
		\vspace{-5pt}
		\caption{\small\textbf{Analysis of key components.} `Colla.' and `Recon.' indicate the attention collaboration module and BEV reconstruction module, respectively.}
		\label{table:key}
		\vspace{-5pt}
	\end{centering}
\end{table}

\begin{table}
	\centering
	\small
	\resizebox{0.49\textwidth}{!}{
		\setlength\tabcolsep{1pt}
		\renewcommand\arraystretch{1.05}
		\begin{tabular}{c||c|cc|c}
			\rowcolor{mygray}
			\thickhline
			& Spatial Compre- & {Default} & {Culver} & {Transmitted  } \\ 
			\rowcolor{mygray}
			\multirow{-2}{*}{Method}  & {ssion Ratio} & AP@$0.7$ & AP@$0.7$ &  Data Size (KB) \\ 
			\hline\hline
			DiscoNet~\cite{li2021learning} & 1.0 & 87.6  & 81.8 &  144.2   \\ \hline
			& 1.0 & 88.6 & 82.8 & 144.2  \\
			& 0.8 (\textit{default}) & 88.3 & 82.6 & 115.3  \\
			& 0.6 & 87.5 & 82.0 & 86.5  \\
			& 0.4 & 87.1 & 81.5 & 57.7  \\
			& 0.2 & 83.0 & 77.6 & 28.8  \\
			\multirow{-6}{*}{\core} & 0.1 & 78.3 & 74.0 & 14.4  \\
			\hline
		\end{tabular}
	}
	\vspace{-5pt}
	\caption{\small\textbf{Analysis of spatial-wise feature compression.} Here compression ratio is defined as the ratio between \textit{uncompressed data size} and \textit{total data size}. In our method, the spatial compression ratio can be calculated as $R\%\!\times\!K\%$.}
	\vspace{-8pt}
	\label{table:compress}
\end{table}

\begin{figure}[t]
	\begin{minipage}[t]{0.5\linewidth}
		\centering
		\includegraphics[width=\textwidth]{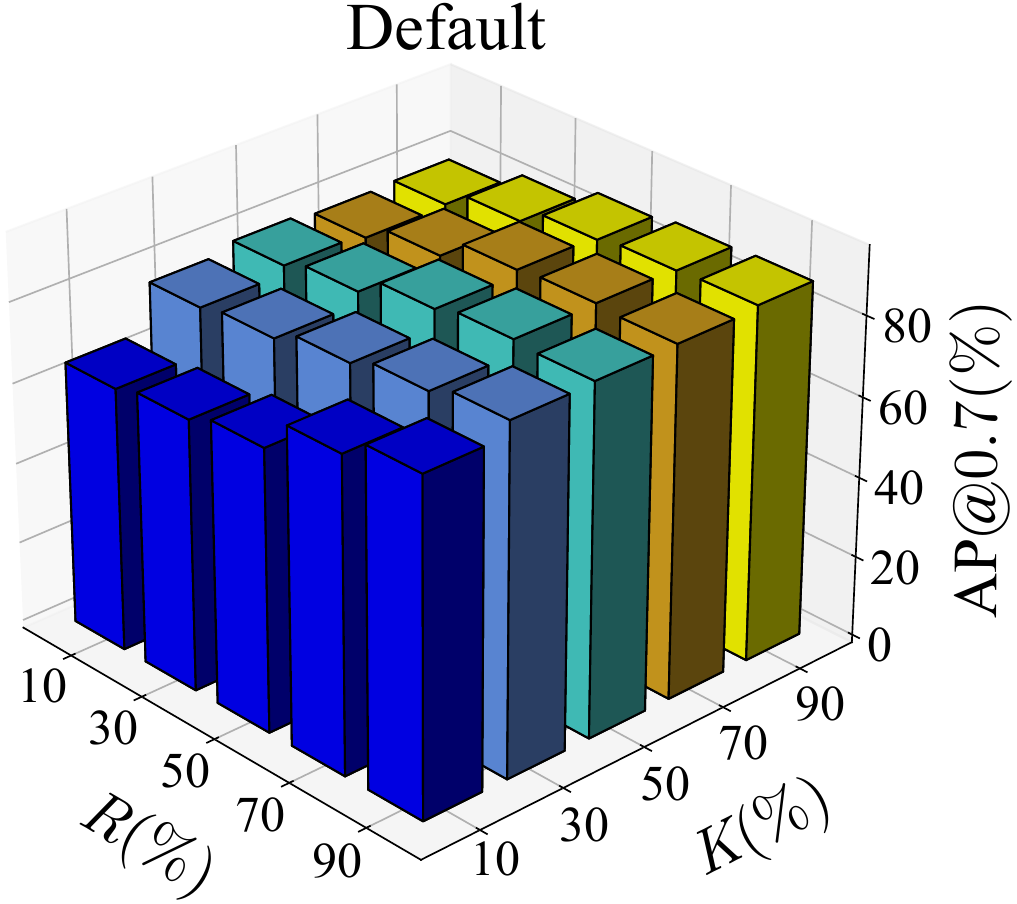}
	\end{minipage}%
	\begin{minipage}[t]{0.5\linewidth}
		\centering
		\includegraphics[width=\textwidth]{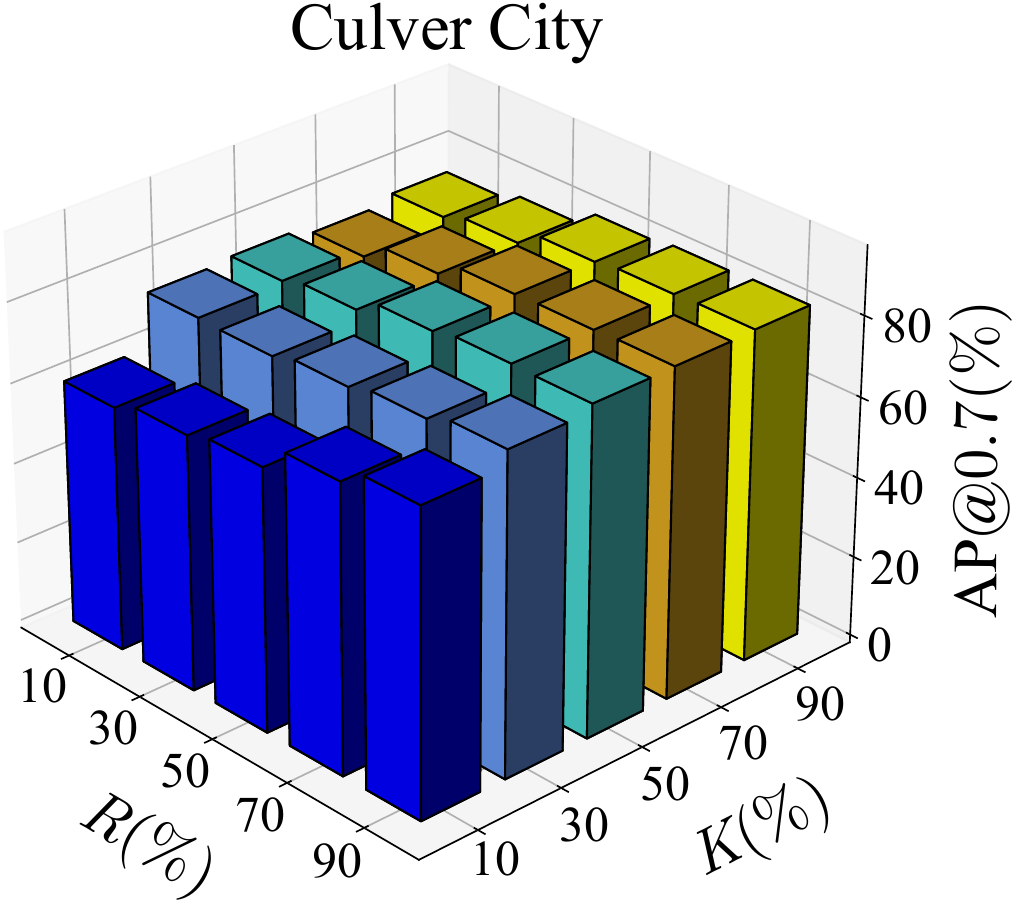}
	\end{minipage}
	\caption{\small\textbf{Hyperparameter analysis} of $K$ and $R$.}
	\label{fig:samp}
\end{figure}

\begin{table}[t]
	\centering
	\small
	\resizebox{0.49\textwidth}{!}{
		\setlength\tabcolsep{8pt}
		\renewcommand\arraystretch{1.0}
		\begin{tabular}{c||cc|cc}
			\rowcolor{mygray}
			\thickhline
			\textbf{$\lambda$} & \multicolumn{2}{c|}{{Default}} & \multicolumn{2}{c}{{Culver City}}\\
			\rowcolor{mygray}
			(Eq.~\ref{eq:loss}) & AP@$0.5$ & AP@$0.7$ & AP@$0.5$ & AP@$0.7$\\ \hline\hline
			$0.1$ & 90.7 & 85.5 & 84.2 & 75.4 \\
			$0.2$ & 90.8 & 86.0 & 84.7 & 76.1 \\
			$0.4$ & 90.5 & 85.9 & 85.3 & 76.8 \\
			$0.6$ & 90.9 & 86.2 & 86.5 & 78.0 \\
			$0.8$ & 90.8 & 86.6 & 86.9 & 78.1 \\
			$\textbf{1.0}$ & \textbf{91.1} & \textbf{86.9} & \textbf{87.2} & \textbf{78.7} \\
			$2.0$ & 89.8 & 83.0 & 85.5 & 77.8 \\ \hline
		\end{tabular}
	}
	\vspace{-5pt}
	\caption{\small\textbf{Hyperparameter analysis} of $\lambda$ in Eq.~\ref{eq:loss}.}
	\vspace{-5pt}
	\label{table:coeff}
\end{table}

\noindent\textbf{Hyperparameters $R$ and $K$.} Next, we study the impacts of hyperparameters $R$ and $K$ (\S\ref{sec:com}), which jointly determine the spatial compression ratio. We vary their values from 10\% to 90\%, and show the object detection results (AP@$0.7$ in subsets of Default and Culver City) in Figure~\ref{fig:samp}. As expected, \core tends to perform worse as $R$ or $K$ becomes smaller. However, we observe that \core is generally robust to these hyperparameters, as even with a compression ratio as low as 10\%, the performance degradation is moderate. These results demonstrate the potential of our \core to be applied to a wide range of scenarios, where different levels of compression ratios may be required due to varying bandwidth and storage constraints.

\noindent\textbf{Hyperparameter $\lambda$.} Table~\ref{table:coeff} measures the effects of the hyperparameter $\lambda$ in Eq.~\ref{eq:loss}. As can be seen, the optimal performance is achieved at  $\lambda=1$, and we observe a slight degradation in performance when $\lambda$ is either increased or decreased from this optimal value.
However, \core exhibits robustness to variations in $\lambda$, as evidenced by the relatively stable results across different values.

\section{Conclusion}
This paper has introduced \core to address cooperative perception in multi-agent scenarios. By solving the task from a  cooperative reconstruction view, \core is able to learn more effective multi-agent collaboration, which is beneficial to specific perception tasks. In addition, cooperative reconstruction  naturally links to the idea of masked data modeling, inspiring us to mask spatial features to further reduce transmitted data volume. \core demonstrates superior performance-bandwidth trade-off on OPV2V in both 3D object detection and BEV semantic segmentation tasks.

{\small
\bibliographystyle{ieee_fullname}
\bibliography{egbib}

\begin{thebibliography}{10}\itemsep=-1pt

\bibitem{arnold2020cooperative}
Eduardo Arnold, Mehrdad Dianati, Robert de Temple, and Saber Fallah.
\newblock Cooperative perception for 3d object detection in driving scenarios
  using infrastructure sensors.
\newblock {\em IEEE Transactions on Intelligent Transportation Systems},
  23(3):1852--1864, 2020.

\bibitem{chen2019f}
Qi Chen, Xu Ma, Sihai Tang, Jingda Guo, Qing Yang, and Song Fu.
\newblock F-cooper: Feature based cooperative perception for autonomous vehicle
  edge computing system using 3d point clouds.
\newblock In {\em Proceedings of ACM/IEEE Symposium on Edge Computing}, 2019.

\bibitem{chen2019cooper}
Qi Chen, Sihai Tang, Qing Yang, and Song Fu.
\newblock Cooper: Cooperative perception for connected autonomous vehicles
  based on 3d point clouds.
\newblock In {\em ICDCS}, 2019.

\bibitem{chen2022model}
Weizhe Chen, Runsheng Xu, Hao Xiang, Lantao Liu, and Jiaqi Ma.
\newblock Model-agnostic multi-agent perception framework.
\newblock {\em arXiv preprint arXiv:2203.13168}, 2022.

\bibitem{cui2022coopernaut}
Jiaxun Cui, Hang Qiu, Dian Chen, Peter Stone, and Yuke Zhu.
\newblock Coopernaut: end-to-end driving with cooperative perception for
  networked vehicles.
\newblock In {\em CVPR}, 2022.

\bibitem{dosovitskiy2017carla}
Alexey Dosovitskiy, German Ros, Felipe Codevilla, Antonio Lopez, and Vladlen
  Koltun.
\newblock Carla: An open urban driving simulator.
\newblock In {\em {CoRL}}, 2017.

\bibitem{feichtenhofermasked}
Christoph Feichtenhofer, Haoqi Fan, Yanghao Li, and Kaiming He.
\newblock Masked autoencoders as spatiotemporal learners.
\newblock In {\em NeurIPS}, 2022.

\bibitem{han2023collaborative}
Yushan Han, Hui Zhang, Huifang Li, Yi Jin, Congyan Lang, and Yidong Li.
\newblock Collaborative perception in autonomous driving: Methods, datasets and
  challenges.
\newblock {\em arXiv preprint arXiv:2301.06262}, 2023.

\bibitem{he2022masked}
Kaiming He, Xinlei Chen, Saining Xie, Yanghao Li, Piotr Doll{\'a}r, and Ross
  Girshick.
\newblock Masked autoencoders are scalable vision learners.
\newblock In {\em CVPR}, 2022.

\bibitem{hu2022where2comm}
Yue Hu, Shaoheng Fang, Zixing Lei, Yiqi Zhong, and Siheng Chen.
\newblock Where2comm: Communication-efficient collaborative perception via
  spatial confidence maps.
\newblock {\em arXiv preprint arXiv:2209.12836}, 2022.

\bibitem{kingma2014adam}
Diederik~P Kingma and Jimmy Ba.
\newblock Adam: A method for stochastic optimization.
\newblock {\em arXiv preprint arXiv:1412.6980}, 2014.

\bibitem{lang2019pointpillars}
Alex~H Lang, Sourabh Vora, Holger Caesar, Lubing Zhou, Jiong Yang, and Oscar
  Beijbom.
\newblock Pointpillars: Fast encoders for object detection from point clouds.
\newblock In {\em CVPR}, 2019.

\bibitem{lei2022latency}
Zixing Lei, Shunli Ren, Yue Hu, Wenjun Zhang, and Siheng Chen.
\newblock Latency-aware collaborative perception.
\newblock In {\em ECCV}, 2022.

\bibitem{Li_2021_RAL}
Yiming Li, Dekun Ma, Ziyan An, Zixun Wang, Yiqi Zhong, Siheng Chen, and Chen
  Feng.
\newblock V2x-sim: A virtual collaborative perception dataset and benchmark for
  autonomous driving.
\newblock 2022.

\bibitem{li2021learning}
Yiming Li, Shunli Ren, Pengxiang Wu, Siheng Chen, Chen Feng, and Wenjun Zhang.
\newblock Learning distilled collaboration graph for multi-agent perception.
\newblock In {\em NeurIPS}, 2021.

\bibitem{li2022multi}
Yiming Li, Juexiao Zhang, Dekun Ma, Yue Wang, and Chen Feng.
\newblock Multi-robot scene completion: Towards task-agnostic collaborative
  perception.
\newblock In {\em {CoRL}}, 2022.

\bibitem{li2020mechanism}
Zhi Li, Ali~Vatankhah Barenji, Jiazhi Jiang, Ray~Y Zhong, and Gangyan Xu.
\newblock A mechanism for scheduling multi robot intelligent warehouse system
  face with dynamic demand.
\newblock {\em Journal of Intelligent Manufacturing}, 31:469--480, 2020.

\bibitem{li2023human}
Zhijun Li, Qinjian Li, Pengbo Huang, Haisheng Xia, and Guoxin Li.
\newblock Human-in-the-loop adaptive control of a soft exo-suit with actuator
  dynamics and ankle impedance adaptation.
\newblock {\em IEEE Transactions on Cybernetics}, 2023.

\bibitem{liu2019fusioneye}
Hansi Liu, Pengfei Ren, Shubham Jain, Mohannad Murad, Marco Gruteser, and Fan
  Bai.
\newblock Fusioneye: Perception sharing for connected vehicles and its
  bandwidth-accuracy trade-offs.
\newblock In {\em IEEE International Conference on Sensing, Communication, and
  Networking (SECON)}, 2019.

\bibitem{liu2020when2com}
Yen-Cheng Liu, Junjiao Tian, Nathaniel Glaser, and Zsolt Kira.
\newblock When2com: Multi-agent perception via communication graph grouping.
\newblock In {\em CVPR}, 2020.

\bibitem{liu2020who2com}
Yen-Cheng Liu, Junjiao Tian, Chih-Yao Ma, Nathan Glaser, Chia-Wen Kuo, and
  Zsolt Kira.
\newblock Who2com: Collaborative perception via learnable handshake
  communication.
\newblock In {\em {ICRA}}, 2020.

\bibitem{long2015fully}
Jonathan Long, Evan Shelhamer, and Trevor Darrell.
\newblock Fully convolutional networks for semantic segmentation.
\newblock In {\em CVPR}, 2015.

\bibitem{meng2021towards}
Qinghao Meng, Wenguan Wang, Tianfei Zhou, Jianbing Shen, Yunde Jia, and Luc
  Van~Gool.
\newblock Towards a weakly supervised framework for 3d point cloud object
  detection and annotation.
\newblock {\em IEEE TPAMI}, 44(8):4454--4468, 2021.

\bibitem{pang2022masked}
Yatian Pang, Wenxiao Wang, Francis~EH Tay, Wei Liu, Yonghong Tian, and Li Yuan.
\newblock Masked autoencoders for point cloud self-supervised learning.
\newblock In {\em ECCV}, 2022.

\bibitem{ren2015faster}
Shaoqing Ren, Kaiming He, Ross Girshick, and Jian Sun.
\newblock Faster r-cnn: Towards real-time object detection with region proposal
  networks.
\newblock In {\em NeurIPS}, 2015.

\bibitem{renrobust}
Shunli Ren, Zixing Lei, Zi Wang, Siheng Chen, and Wenjun Zhang.
\newblock Robust collaborative perception against communication interruption.

\bibitem{xu2022cobevt}
Hao Xiang Wei Shao Bolei Zhou Jiaqi~Ma Runsheng~Xu, Zhengzhong~Tu.
\newblock Cobevt: Cooperative bird's eye view semantic segmentation with sparse
  transformers.
\newblock In {\em {CoRL}}, 2022.

\bibitem{sarkar2018scalable}
Chayan Sarkar, Himadri~Sekhar Paul, and Arindam Pal.
\newblock A scalable multi-robot task allocation algorithm.
\newblock In {\em {ICRA}}, 2018.

\bibitem{scherer2015autonomous}
J{\"u}rgen Scherer, Saeed Yahyanejad, Samira Hayat, Evsen Yanmaz, Torsten
  Andre, Asif Khan, Vladimir Vukadinovic, Christian Bettstetter, Hermann
  Hellwagner, and Bernhard Rinner.
\newblock An autonomous multi-uav system for search and rescue.
\newblock In {\em Proceedings of the First Workshop on Micro Aerial Vehicle
  Networks, Systems, and Applications for Civilian Use}, 2015.

\bibitem{su2022uncertainty}
Sanbao Su, Yiming Li, Sihong He, Songyang Han, Chen Feng, Caiwen Ding, and Fei
  Miao.
\newblock Uncertainty quantification of collaborative detection for
  self-driving.
\newblock {\em arXiv preprint arXiv:2209.08162}, 2022.

\bibitem{vadivelu2021learning}
Nicholas Vadivelu, Mengye Ren, James Tu, Jingkang Wang, and Raquel Urtasun.
\newblock Learning to communicate and correct pose errors.
\newblock In {\em {CoRL}}, 2021.

\bibitem{volk2019environment}
Georg Volk, Alexander von Bernuth, and Oliver Bringmann.
\newblock Environment-aware development of robust vision-based cooperative
  perception systems.
\newblock In {\em IEEE Intelligent Vehicles Symposium (IV)}, 2019.

\bibitem{wang2020v2vnet}
Tsun-Hsuan Wang, Sivabalan Manivasagam, Ming Liang, Bin Yang, Wenyuan Zeng, and
  Raquel Urtasun.
\newblock V2vnet: Vehicle-to-vehicle communication for joint perception and
  prediction.
\newblock In {\em ECCV}, 2020.

\bibitem{wang2021exploring}
Wenguan Wang, Tianfei Zhou, Fisher Yu, Jifeng Dai, Ender Konukoglu, and Luc
  Van~Gool.
\newblock Exploring cross-image pixel contrast for semantic segmentation.
\newblock In {\em ICCV}, 2021.

\bibitem{wu2013online}
Yi Wu, Jongwoo Lim, and Ming-Hsuan Yang.
\newblock Online object tracking: A benchmark.
\newblock In {\em CVPR}, 2013.

\bibitem{xu2021opencda}
Runsheng Xu, Yi Guo, Xu Han, Xin Xia, Hao Xiang, and Jiaqi Ma.
\newblock Opencda: an open cooperative driving automation framework integrated
  with co-simulation.
\newblock In {\em ITSC}, 2021.

\bibitem{xu2022v2x}
Runsheng Xu, Hao Xiang, Zhengzhong Tu, Xin Xia, Ming-Hsuan Yang, and Jiaqi Ma.
\newblock V2x-vit: Vehicle-to-everything cooperative perception with vision
  transformer.
\newblock In {\em ECCV}, 2022.

\bibitem{xu2022opv2v}
Runsheng Xu, Hao Xiang, Xin Xia, Xu Han, Jinlong Li, and Jiaqi Ma.
\newblock Opv2v: An open benchmark dataset and fusion pipeline for perception
  with vehicle-to-vehicle communication.
\newblock In {\em {ICRA}}, 2022.

\bibitem{yu2022dair}
Haibao Yu, Yizhen Luo, Mao Shu, Yiyi Huo, Zebang Yang, Yifeng Shi, Zhenglong
  Guo, Hanyu Li, Xing Hu, Jirui Yuan, et~al.
\newblock Dair-v2x: A large-scale dataset for vehicle-infrastructure
  cooperative 3d object detection.
\newblock In {\em CVPR}, 2022.

\bibitem{yuan2022keypoints}
Yunshuang Yuan, Hao Cheng, and Monika Sester.
\newblock Keypoints-based deep feature fusion for cooperative vehicle detection
  of autonomous driving.
\newblock {\em IEEE Robotics and Automation Letters}, 7(2):3054--3061, 2022.

\bibitem{zaccaria2021multi}
Michela Zaccaria, Mikhail Giorgini, Riccardo Monica, and Jacopo Aleotti.
\newblock Multi-robot multiple camera people detection and tracking in
  automated warehouses.
\newblock In {\em International Conference on Industrial Informatics}, 2021.

\bibitem{zhou2022survey}
Tianfei Zhou, Fatih Porikli, David~J Crandall, Luc Van~Gool, and Wenguan Wang.
\newblock A survey on deep learning technique for video segmentation.
\newblock {\em IEEE TPAMI}, 45(6):7099--7122, 2022.

\bibitem{zhou2022rethinking}
Tianfei Zhou, Wenguan Wang, Ender Konukoglu, and Luc Van~Gool.
\newblock Rethinking semantic segmentation: A prototype view.
\newblock In {\em CVPR}, 2022.

\bibitem{zhou2018voxelnet}
Yin Zhou and Oncel Tuzel.
\newblock Voxelnet: End-to-end learning for point cloud based 3d object
  detection.
\newblock In {\em CVPR}, 2018.

\bibitem{zhou2022multi}
Yang Zhou, Jiuhong Xiao, Yue Zhou, and Giuseppe Loianno.
\newblock Multi-robot collaborative perception with graph neural networks.
\newblock {\em IEEE Robotics and Automation Letters}, 7(2):2289--2296, 2022.

\end{thebibliography}
}

\clearpage

\appendix
\setcounter{table}{0}
\renewcommand{\thetable}{A\arabic{table}}
\setcounter{figure}{0}
\renewcommand{\thefigure}{A\arabic{figure}}

\section{More Experimental Result on V2X-Sim~\cite{Li_2021_RAL}}

Table~\ref{tab:v2x} presents the quantitative results on the V2X-Sim~\cite{Li_2021_RAL} dataset. A single-agent perception model that operates on a single-view point cloud without collaboration represents the \textit{Lower-Bound}, whereas an early collaboration model that communicates raw point cloud data directly represents the \textit{Upper-Bound}. As observed, for object detection, \core achieves APs of \textbf{70.0\%} and \textbf{64.8\%} at $0.5$ and $0.7$ IoU thresholds, which are the highest among all methods. For vehicle segmentation, \core reaches a \textbf{58.3\%} score, higher than most previous works, except for \textit{V2VNet}~\cite{wang2020v2vnet} which has 58.4\%. In terms of mIoU, \textit{V2VNet}~\cite{wang2020v2vnet} achieves the best performance at 41.1\%, while \core is competitive at \textbf{40.5\%}. Despite this, \core still significantly outperforms other notable models such as \textit{When2com}~\cite{liu2020when2com}, \textit{Who2com}~\cite{liu2020who2com}, and \textit{DiscoNet}~\cite{li2021learning}. More importantly, the performance gap between \core and the \textit{Upper-Bound} is significantly reduced compared to previous methods.

\vspace{-5pt}
\section{More Insights on Reconstruction}

\subsection{Supervisory BEV Generation}

Given an ideal 3D point cloud $\tilde{\bm{S}}_i$ (\S\ref{sec:rec}), we discuss two supervisory BEV generation methods: CNN-based or grid-based projection. The former adopts existing 3D detectors (e.g., VoxelNet, PointPillar) to generate BEV features. Taking VoxelNet as an example backbone, it first voxelizes $\tilde{\bm{S}}_i$ into a 3D voxel grid and then extracts voxel features using PointNet. Subsequently, a series of 3D convolutions are applied to further aggregate voxel features, thus generating the BEV representation of the 3D scene. Grid-based projection directly maps points to a two-dimensional grid, where each grid cell represents a region, and the presence or absence of points is used to indicate whether that region contains point cloud data. The resulting BEV map is a two-dimensional array, where each element represents a region, reflecting the distribution of the original point cloud on the BEV plane based on the arrangement of points.

Table~\ref{table:bev} presents the comparison results of generating supervisory BEV using the two methods. As seen, both methods can generate promising supervisory BEV, but the CNN-based projection appears to have a slight advantage, especially in more complex environments (\ie, Culver City). This indicates that the features extracted by the CNN-based method encode richer point cloud structural information, resulting in a more semantically meaningful BEV representation, while the grid-based method may suffer some loss during the BEV generation process.

\subsection{Loss Curve}
\begin{figure}[t]
	\begin{center}
		\includegraphics[width=\linewidth]{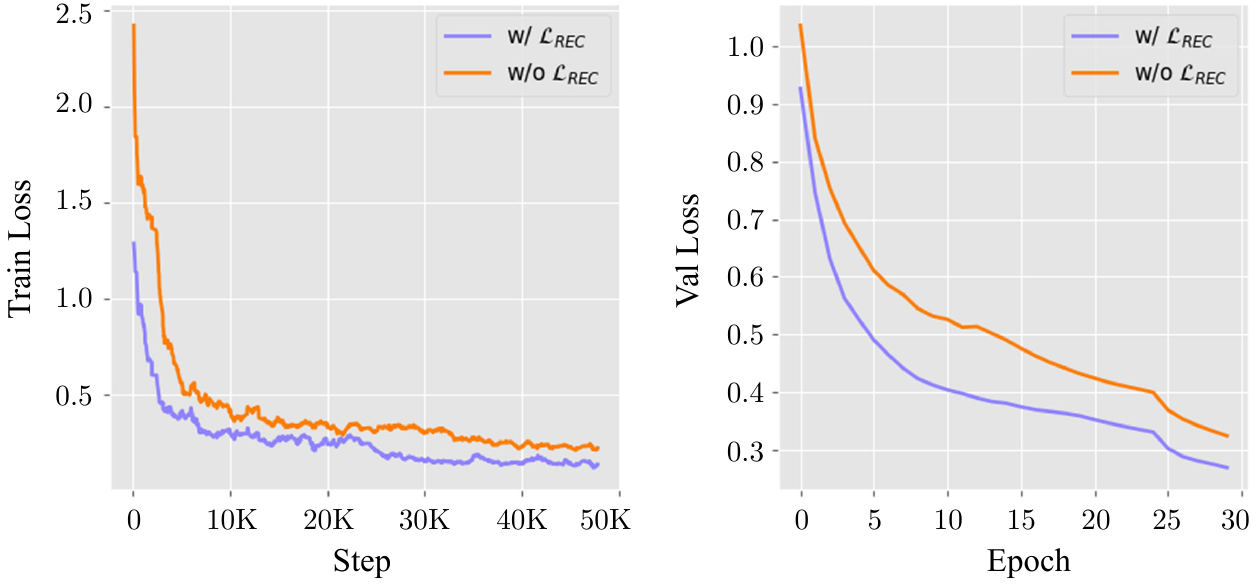}
	\end{center}
	\vspace{-18pt}
	\caption{\small{Loss curves of \core} trained w/ or w/o $\mathcal{L}_\text{REC}$.}
	\label{fig:loss_curve}
		\vspace{-10pt}
\end{figure}

To gain more insights into the role of reconstruction, we show loss curves of \core that is trained w/ and w/o $\mathcal{L}_\text{REC}$ in Fig.~\ref{fig:loss_curve}. It can be observed that when utilizing $\mathcal{L}_\text{REC}$, the model exhibits convergence towards a more optimal solution. The addition of the reconstruction loss term $\mathcal{L}_\text{REC}$ encourages the model to effectively reconstruct the input data, thereby capturing meaningful features and reducing information loss. As a result, the training loss steadily decreases over iterations, indicating that the model is learning to better represent the underlying patterns and structures in the data. In contrast, the model trained without $\mathcal{L}_\text{REC}$ shows a slower convergence rate and a larger final training loss, suggesting a less effective representation learning process. These findings highlight the importance of incorporating $\mathcal{L}_\text{REC}$ in the training process, as it enhances the model's ability to uncover and utilize relevant features, leading to improved performance and generalization capabilities.

\begin{table}[t]
	\centering
	\resizebox{0.49\textwidth}{!}{
		\setlength\tabcolsep{10pt}
		\renewcommand\arraystretch{1.0}
		\begin{tabular}{r||cc|cc}
			\rowcolor{mygray}
			\thickhline
			\multirow{2}{*}{} & \multicolumn{2}{c|}{{Object Detection}} & \multicolumn{2}{c}{{Segmentation}}\\
			\rowcolor{mygray}
			\multirow{-2}{*}{{Method}} & AP@$0.5$ & AP@$0.7$ & Vehicle & mIoU\\ \hline\hline
			Lower-Bound  & 49.9 & 44.2 & 45.9 & 36.6 \\
			Upper-Bound & 70.4 & 67.0 & 64.1 & 42.3 \\\hline
			When2com\!~\cite{liu2020when2com} & 44.0 & 39.9 & 48.4 & 34.5  \\
			Who2com\!~\cite{liu2020who2com} & 44.0 & 39.9 & 48.4 & 34.3 \\
			V2VNet\!~\cite{wang2020v2vnet} & 68.4 & 62.8 & \textbf{58.4} & \textbf{41.1} \\
			DiscoNet\!~\cite{li2021learning} & 69.0 & 63.4 & 56.7 & 40.8 \\
			\core & \textbf{70.0} & \textbf{64.8} & 58.3 & 40.5 \\\hline
		\end{tabular}
	}
	\vspace{-5pt}
	\caption{\small{Quantitative results} on V2X-Sim~\cite{Li_2021_RAL}.}
		\vspace{-5pt}
	\label{tab:v2x}
\end{table}

\begin{table}[t]
	\centering
	\small
	\resizebox{0.49\textwidth}{!}{
		\setlength\tabcolsep{6pt}
		\renewcommand\arraystretch{1.0}
		\begin{tabular}{c||cc|cc}
			\rowcolor{mygray}
			\thickhline
			\multirow{2}{*}{} & \multicolumn{2}{c|}{{Default}} & \multicolumn{2}{c}{{Culver City}}\\
			\rowcolor{mygray}
			\multirow{-2}{*}{{Method}} & AP@$0.5$ & AP@$0.7$ & AP@$0.5$ & AP@$0.7$\\ \hline\hline
			Grid-based & 90.2 & {88.3} & 87.0 & 81.6 \\
			CNN-based & {90.9} & {88.3} & {87.8} & {82.6} \\\hline
		\end{tabular}
	}
	\vspace{-5pt}
	\caption{\small Results of different ways to create  supervisory BEV.}
	\vspace{-5pt}
	\label{table:bev}
\end{table}

\subsection{BEV Visualization}
\begin{figure*}[t]
	\begin{center}
		\includegraphics[width=\linewidth]{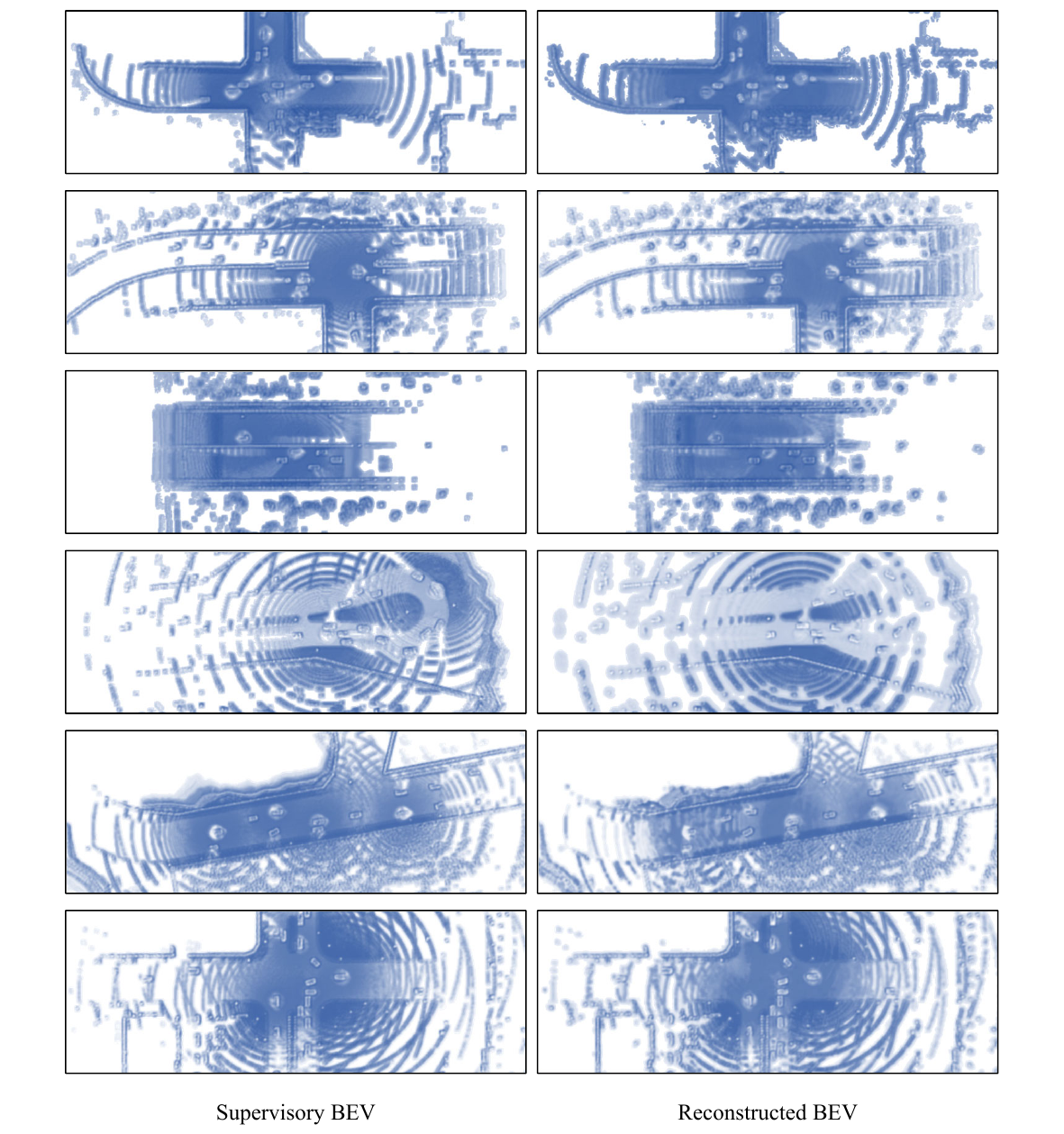}
	\end{center}
	\vspace{-15pt}
	\caption{\small\textbf{Visualization} of the reconstructed BEV features.}
	\label{fig:vis_rec}
\end{figure*}

Fig.~\ref{fig:vis_rec} depicts  reconstructed BEV features, showcasing the high fidelity achieved in capturing the details of the supervisory BEV. As seen, the reconstructed BEV (left) closely resemble the supervisory BEV (right), confirming the effectiveness of the reconstruction process.

\section{Additional Qualitative Result}
\begin{figure*}[t]
	\begin{center}
		\includegraphics[width=\linewidth]{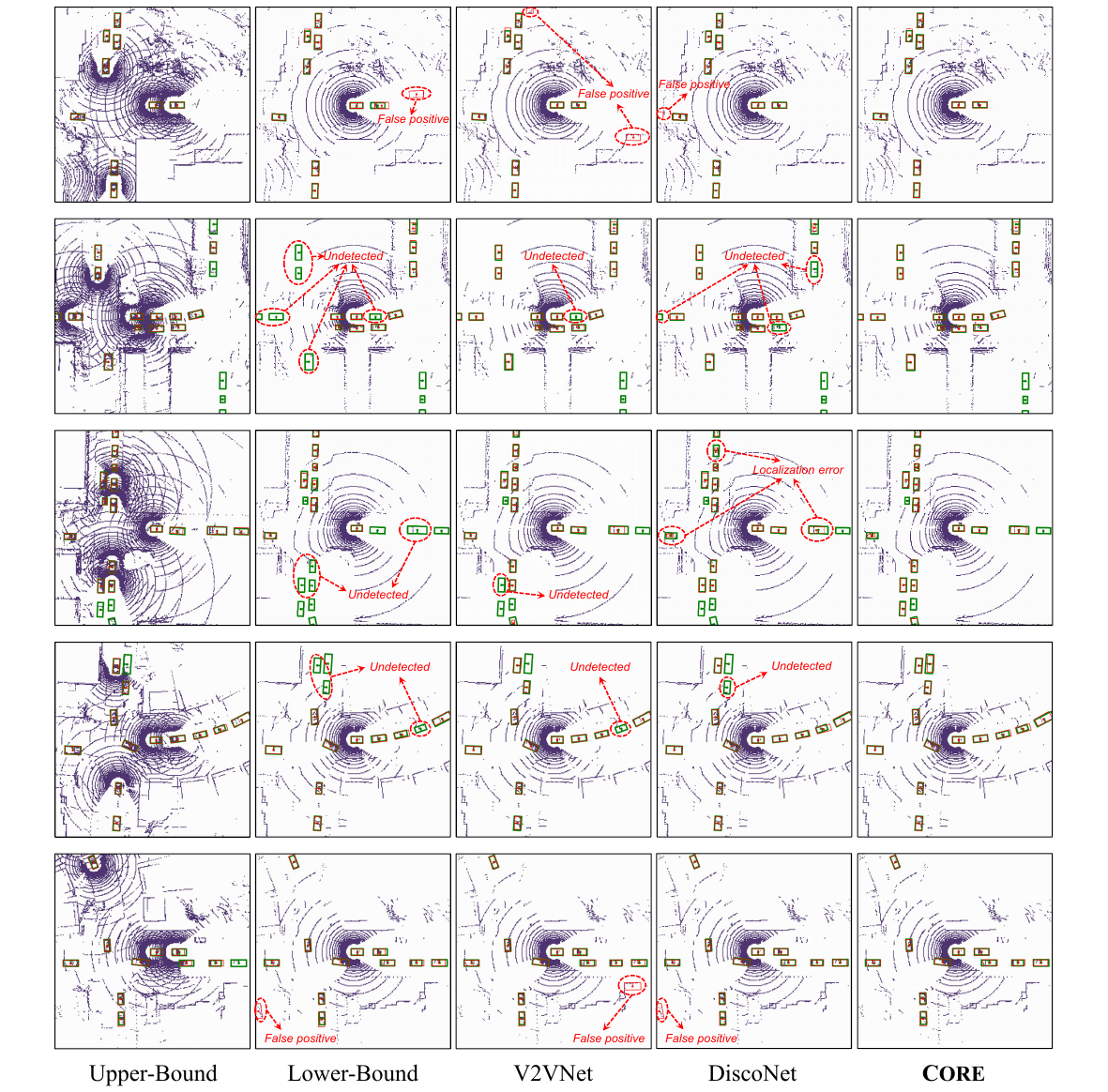}
	\end{center}
	\vspace{-15pt}
	\caption{\small\textbf{Qualitative object detection results} on V2X-Sim~\cite{Li_2021_RAL}.}
	\label{fig:v2x_det}
\end{figure*}

\begin{figure*}[t]
	\begin{center}
		\includegraphics[width=\linewidth]{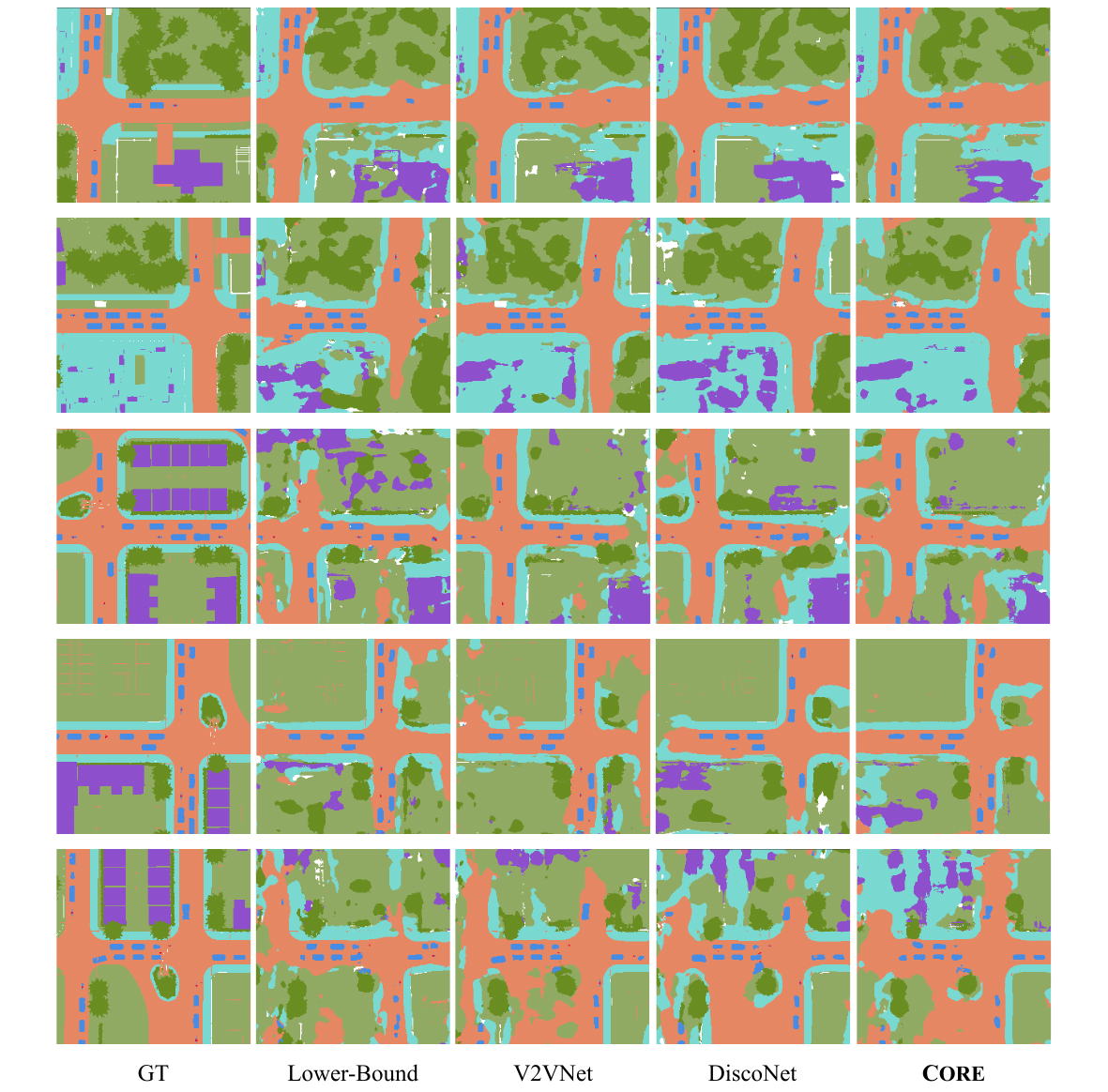}
	\end{center}
	\vspace{-15pt}
	\caption{\small\textbf{Qualitative semantic segmentation results} on V2X-Sim~\cite{Li_2021_RAL}.}
	\label{fig:v2x_seg}
\end{figure*}

We show additional qualitative results on V2X-Sim~\cite{Li_2021_RAL} (Fig.~\ref{fig:v2x_det} and Fig.~\ref{fig:v2x_seg}). \core demonstrates strong performance in both object detection and semantic segmentation.

\clearpage

\end{document}